\documentclass[runningheads]{llncs}
 
 \usepackage{eccv}

\usepackage{threeparttable}
\RequirePackage[textwidth=14.0cm,textheight=20.0cm,centering]{geometry}

\usepackage{eccvabbrv}

\usepackage{graphicx}
\usepackage{booktabs}

\usepackage[accsupp]{axessibility}  

\usepackage[pagebackref,breaklinks,colorlinks,citecolor=eccvblue]{hyperref}

\usepackage{orcidlink}

\begin{document}


\title{\large ProtoPathway: Biologically Structured Prototype-Pathway Fusion for Multimodal Cancer Survival Prediction}

\titlerunning{ProtoPathway}


\author{Amaya Gallagher-Syed\inst{1,2,}\thanks{Corresponding author. \\ Under review.}\and
Costantino Pitzalis\inst{1} \and
Myles J. Lewis \inst{1} \and
Michael R. Barnes \inst{1} \and
Gregory Slabaugh \inst{1} }

\authorrunning{Gallagher-Syed et al.}

\institute{Queen Mary University of London \and
Imperial College London \\
\email{\{a.r.syed, c.pitzalis, myles.lewis, m.r.barnes, g.slabaugh\}@qmul.ac.uk} \\
\email{a.gallaghersyed@imperial.ac.uk}
}

\maketitle


\begin{abstract}
We introduce ProtoPathway, an interpretable-by-design multimodal framework for cancer survival prediction that unifies whole slide imaging and transcriptomics through encoders producing biologically grounded representations on both sides of the fusion. On the histopathology side, $K$ learnable morphological prototypes, trained end-to-end with the survival objective, serve as the slide representation itself: patches flow into prototype tokens via soft assignment, compressing variable-length patch sets into fixed task-adaptive tokens. On the genomic side, a bipartite graph neural network encodes gene expression within the Reactome pathway hierarchy, producing pathway embeddings that reflect both constituent genes and their broader biological context through bidirectional message passing over a shared gene--pathway graph. Cross-modal attention then operates over a compact prototype $\times$ pathway matrix in which prototypes query pathways, modeling the biological direction in which molecular programs give rise to tissue morphology. Because both axes carry stable task-learned identity, the attention matrix is itself an interpretability output, yielding native inference-time attribution across the full biological hierarchy, from genes through pathways and prototypes to spatial tissue maps. We evaluate on five TCGA cancer cohorts, demonstrating competitive or superior survival prediction with substantially improved biological interpretability and reduced computational cost, with interpretability claims validated through fold-stratified rank-based population-level analysis. Our source code, model weights, and Reactome pathways, together with a unified codebase reimplementing all multimodal survival baselines under identical preprocessing and evaluation, are available at: \href{https://github.com/AmayaGS/ProtoPathway}{\texttt{https://github.com/AmayaGS/ProtoPathway}}.

\end{abstract}


\section{Introduction}
\label{sec:intro}

Whole slide images (WSIs) and bulk transcriptomic profiles offer complementary views of tumor biology: gene expression captures the active biological programs governing disease progression, while histopathological images capture the spatial organization and cellular morphology resulting from those programs~\cite{wang2021,Lipkova2022}. Their integration for survival prediction has attracted sustained attention, motivated by the prospect of more accurate prognosis and deeper insight into how molecular states manifest as observable tissue phenotypes~\cite{Chen2022_pathomic}. Yet for clinical adoption, prediction accuracy alone is insufficient. Clinicians require models whose reasoning can be validated against established biology, particularly when predictions inform treatment decisions with direct consequences for patient outcomes~\cite{Rudin2019}. This has motivated substantial work on post-hoc explainability methods such as SHAP~\cite{shap} and LIME~\cite{lime}, which treat trained models as black boxes and approximate their behavior through external perturbation or decomposition. However, such methods offer no guarantees that explanations reflect actual model behavior~\cite{Slack2020,Chattopadhyay2023}, and in multimodal settings typically explain each modality in isolation, without revealing the cross-modal relationships that are often most scientifically valuable.

\par\vspace{3pt}

\noindent These limitations motivate designing architectures that are interpretable by design, where predictions flow through representations with inherent semantic meaning rather than being explained retrospectively. Here, a model's reasoning is transparent not because an external method has approximated it, but because the representations themselves carry domain-grounded identity and interpretability is a structural property of the computation~\cite{Kim2017,Yeh2020,Koh2020}. For multimodal survival prediction, this places a specific requirement on cross-modal fusion: the representations being fused must carry stable, semantically grounded identity on both sides. Hence, the encoders on both sides need to align with biologically grounded meaning.

\subsection{Related Work}
\label{sec:related}

The encoders used in multimodal survival prediction are drawn from established unimodal traditions. We therefore briefly review encoders for each modality, before turning to multimodal fusion methods in survival prediction. 

\par\vspace{3pt}

\noindent\textbf{WSI Encoders.} On the WSI side, multiple instance learning (MIL) over patch features is the dominant paradigm. Attention-based MIL methods such as ABMIL~\cite{abmil}, TransMIL~\cite{transmil}, and DSMIL~\cite{dsmil} aggregate patches into a slide representation via learned attention weights, with no explicit notion of morphological pattern beyond per-patch attention scores. Prototype-based MIL methods take a different approach, introducing prototype vectors that capture recurring morphological patterns. How these prototypes are obtained and used varies considerably. ProtoMIL~\cite{protomil}, inheriting from ProtoPNet~\cite{protopnet}, scores patches against learned prototypes and treats the resulting similarity vectors as classifier inputs. TPMIL~\cite{tpmil} adds prototypes as auxiliary refiners alongside an ABMIL classifier that performs the prediction. PANTHER~\cite{panther} fits prototypes via Gaussian mixture models (GMMs) and freezes them as a fixed lookup table. Slot-MIL~\cite{slotmil} adapts slot attention~\cite{slotattention} to MIL, where a fixed number of slots compete via iterative cross-attention to claim patches. The competition runs over slots, so each patient produces its own slot identities with no guarantee of consistency across patients. Across these methods, prototypes function as similarity references, refiners, fixed unsupervised summaries, or per-sample attention outputs rather than as a slide representation with stable identity across patients.

\par\vspace{3pt}

\noindent\textbf{Transcriptomic Encoders.} On the genomic side, the dominant unimodal encoder is the self-normalizing network (SNN)~\cite{snn}, an MLP variant operating on the full gene expression vector without explicit biological structure. Methods that introduce biological structure take several forms. P-NET~\cite{pnet} encodes gene expression through a sparse hierarchical MLP where layer connectivity encodes the biological programs (or pathways) defined in the Reactome database~\cite{reactome}, with no graph structure between pathway nodes. GraphPath~\cite{graphpath} applies multi-head self-attention over a pathway-pathway interaction network, with no gene-level nodes in the graph. Pathformer~\cite{pathformer} combines a pathway-based sparse neural network with a transformer whose attention bias is computed from a pathway-pathway crosstalk network, modeling pathway interactions through attention rather than graph topology. Across these approaches, the biological topology connecting genes and pathways is encoded as architectural sparsity, pathway-level adjacency, or attention bias, but genes and pathways are never modeled as nodes in a shared graph.

\par\vspace{3pt}

\noindent\textbf{Multimodal Fusion.} Multimodal fusion methods inherit from both traditions but face an additional constraint: the encoders on each side determine what semantic identity is available to the cross-modal fusion. PORPOISE~\cite{porpoise} established the paradigm by pairing an ABMIL~\cite{abmil} WSI encoder with a single SNN over the full gene expression vector, fused through Kronecker product interactions. This bilinear late fusion discards biological organization on the transcriptomic side and fails to provide cross-modal attention weights for interpretation. MCAT~\cite{mcat} and MOTCAT~\cite{motcat} introduced cross-modal reasoning by partitioning genes into six coarse functional families, each embedded through a separate SNN, and computing co-attention (MCAT) or optimal transport matching (MOTCAT) between family tokens and individual patch tokens. This enabled richer interaction than late fusion, but the coarse genomic grouping limits biological resolution and the attention matrices lack stable semantic identity on the WSI side, where individual patches do not carry consistent meaning across patients.

\par\vspace{3pt}

\noindent SurvPath~\cite{survpath} substantially increased biological resolution by embedding genes at the level of individual curated pathways, each through an independent SNN, with masked self-attention over the concatenated multimodal token sequence. PIBD~\cite{pibd} adopted a similar pathway vocabulary and introduced prototypical information bottleneck selection to identify discriminative instances, but the prototypes serve as selection filters rather than as semantically grounded representations. Both methods improved genomic granularity while retaining independent per-pathway encoding and token-level fusion that scales with retained patch count. MMP~\cite{mmp} introduces stable identity on the WSI side by compressing patches through PANTHER's~\cite{panther} GMM prototypes and applying cross-attention between these summaries and pathway tokens. However, the GMM prototypes are fit once in preprocessing and frozen, capturing statistically dominant morphological variation with no guarantee of alignment with prognostically relevant features. On the genomic side, MMP retains independent pathway SNNs over a restricted pathway vocabulary.

\par\vspace{3pt}

\noindent Across this progression, no existing multimodal survival prediction method places stable task-learned semantic identity on both axes of the cross-modal attention matrix. On the morphological side, methods either attend over individual patches that lack consistent meaning across patients, or treat prototypes as auxiliary structures rather than as the slide representation itself. In no case do learned task-adaptive tokens serve as both the slide representation and as queries in cross-modal attention. On the genomic side, pathway-level encoding is uniformly implemented through independent SNNs per pathway, so a gene appearing in multiple pathways receives separate representations with no parameter sharing or cross-pathway communication. The biological topology connecting genes to pathways partitions the input, rather than being encoded within the computational graph.


\subsection{Contributions}
\label{sec:intro_pp}

We introduce ProtoPathway, an interpretable-by-design multimodal architecture that compresses tissue morphology into learnable prototypes and encodes pathway topology via bipartite graph neural networks (GNN), enabling cross-modal fusion where every intermediate representation has stable semantic, biologically grounded identity. ProtoPathway makes three contributions:

\begin{enumerate}
    \item \textbf{Learned morphological prototypes as slide representation.} $K$ learnable prototype vectors, trained end-to-end with the survival objective, serve as the slide representation itself rather than as reference points against which patches are scored. Patches flow into prototype tokens via soft assignment, and these tokens feed every downstream operation, enabling their dual role as morphological aggregators and as queries into the pathway space.
    
    \item \textbf{Topology-aware transcriptomic encoding via bidirectional message passing.} A bipartite GNN encodes gene expression within the Reactome pathway hierarchy, where genes and pathways form two node types connected by bidirectional membership edges. Message passing over this topology enables parameter sharing and implicit cross-pathway communication through genes participating in multiple processes, replacing the independent per-pathway SNNs used by existing methods.

    \item \textbf{Semantically grounded asymmetric cross-modal fusion.} Cross-attention operates over a compact prototype $\times$ pathway matrix in which prototypes query pathways, consistent with biology's causal direction from molecular programs to morphological manifestation. Because both axes carry stable task-learned identity, the attention matrix is an interpretability output rather than an internal computation, yielding a traceable link from genes through pathways and prototypes to spatial tissue maps.
\end{enumerate}

\noindent ProtoPathway achieves these properties while maintaining computational efficiency: attention over $K$ prototypes and $P$ pathways replaces token-level attention over thousands of individual patches required by other methods. We evaluate on five TCGA cancer cohorts, demonstrating competitive survival prediction with substantially improved biological interpretability and reduced computational cost.


\section{Methods}
\label{sec:Methods}

\subsection{Architecture Overview}

\begin{figure}[t]
    \makebox[\textwidth][c]{\includegraphics[width=1.15\textwidth]{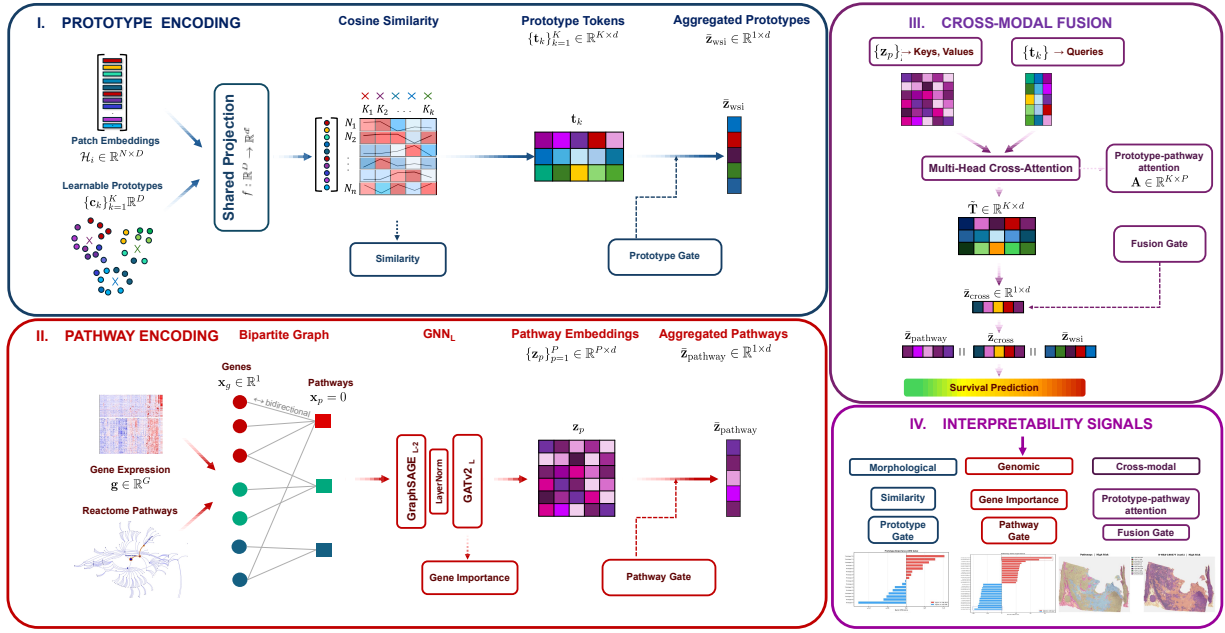}}
    \caption[ProtoPathway architecture]{\textbf{ProtoPathway architecture.} \textbf{ I. Prototype Encoding}: WSI patches are compressed into $K$ learnable prototype tokens $\mathbf{T}$. \textbf{II. Pathway Encoding}: Gene expression is encoded via a bipartite GNN over the Reactome gene--pathway hierarchy, producing pathway embeddings $\mathbf{Z}$. \textbf{III. Cross-Modal Fusion}: Prototypes query pathway embeddings via cross-modal attention, yielding an interpretable $K \times P$ association matrix $\mathbf{A}$. The final representation concatenates genomic, morphological, and cross-modal streams. \textbf{IV. Interpretability Signals}: The pipeline is designed to output interpretability signals directly from inference without post-hoc methods, from genes through pathways and prototypes, to spatial tissue maps.} 
    \label{protopathway_pipeline}
\end{figure}

ProtoPathway consists of three components (Fig.~\ref{protopathway_pipeline}). A learnable prototype module compresses variable-length WSI patch sets into a fixed set of morphological token embeddings (Section~\ref{sec:wsi_encoder}). A bipartite GNN encodes gene expression within the Reactome pathway hierarchy, producing pathway embeddings and patient-level genomic representations (Section~\ref{sec:gene_encoder}). Cross-modal attention fuses the two by letting prototype tokens query pathway embeddings, producing an interpretable association matrix alongside the final patient-level survival representation (Section~\ref{sec:fusion}). The interpretability signals emerging from each component are summarized in Section~\ref{sec:interpretability}.

\subsection{Prototype Encoding}
\label{sec:wsi_encoder}

Tissue regions in gigapixel WSIs are divided into non-overlapping patches and encoded by a frozen pathology foundation model, yielding a bag of patch-level feature vectors $\mathcal{H}_i = \{\mathbf{h}_1, \ldots, \mathbf{h}_{N_i}\}$, $\mathbf{h}_n \in \mathbb{R}^{D}$, for each patient. Although this representation is variable-length and high-dimensional, the underlying morphological landscape is far more structured: tissue organizes into recurring patterns that appear across patients in different spatial configurations. We exploit this by learning $K$ morphological prototypes that decompose the patch distribution into interpretable token embeddings, each capturing a coherent tissue pattern relevant to prognosis.

\par\vspace{3pt}

\noindent\textbf{Prototype Definition and Projection.} We define $K$ learnable prototype vectors $\{\mathbf{c}_k\}_{k=1}^K$ in the input feature space $\mathbb{R}^D$, initialized from $k$-means centroids (\textit{cf.} Supplementary Materials (SM)~\ref{supp:kmeans}). Both patch embeddings and prototype vectors are projected to a shared $d$-dimensional space through a linear transformation $f: \mathbb{R}^{D} \to \mathbb{R}^{d}$, aligning morphological and pathway representations for cross-modal attention.

\par\vspace{3pt}

\noindent\textbf{Soft Prototype Assignment.} Assignment probabilities are computed via temperature scaled cosine similarity between projected patches and prototypes:
\begin{equation}
    \alpha_{nk} = \operatorname{softmax}_k \left( \frac{1}{\tau} \cdot \cos\bigl(f(\mathbf{h}_n),\, f(\mathbf{c}_k)\bigr) \right)
    \label{eq:cosine_sim}
\end{equation}

where $\cos(\cdot,\cdot)$ denotes cosine similarity and $\tau$ controls assignment sharpness. Cosine similarity is robust to staining intensity variation across slides, and soft assignment ($\sum_{k=1}^{K} \alpha_{nk} = 1$) captures the continuous morphological variation characteristic of biological tissues while maintaining differentiability.

\par\vspace{3pt}

\noindent\textbf{Prototype Tokens and Gating.} Each prototype aggregates its assigned patches into token embeddings $\mathbf{t}_k$, and a learnable gating network $w_k$ computes prototype-specific importance scores. The patient-level WSI embedding is the importance-weighted aggregation  $\bar{\mathbf{z}}_{\mathrm{wsi}}$:
\begin{equation}
    \mathbf{t}_k = \frac{\sum_{n=1}^{N} \alpha_{nk}\, f(\mathbf{h}_n)}{\sum_{n=1}^{N} \alpha_{nk}}, \qquad w_k = \operatorname{softmax}_k\bigl(\psi(\mathbf{t}_k)\bigr), \qquad \bar{\mathbf{z}}_{\mathrm{wsi}} = \sum_{k=1}^{K} w_k\, \mathbf{t}_k
\end{equation}
The resulting tokens $\{\mathbf{t}_k\}_{k=1}^K \in \mathbb{R}^{K \times d}$ compress the variable-length patch bag into a fixed set of morphological embeddings. Because prototypes and the survival objective are trained end-to-end, the decomposition captures prognostically relevant variation rather than statistically dominant but potentially uninformative structure. Both the token embeddings $\{\mathbf{t}_k\}_{k=1}^{K}$ and the aggregated $\bar{\mathbf{z}}_{\mathrm{wsi}}$ are passed forward: the former participate in cross-modal attention (Section~\ref{sec:fusion}), the latter contributes to the final fused representation.

\par\vspace{3pt}

\noindent\textbf{Spatial Interpretability.} Since each patch has known tissue coordinates, the assignment matrix maps learned prototypes to spatial slide locations. Hard assignment $k^*_n = \operatorname{argmax}_k\, \alpha_{nk}$ produces per-patch prototype labels that can be projected as spatial overlays, revealing where each morphological pattern is localized. Combined with cross-modal attention (Section~\ref{sec:fusion}), this enables traceability from molecular programs to specific tissue regions. Details are in SM~\ref{sec:sm-spatial-overlays}.

\subsection{Pathway Encoding}
\label{sec:gene_encoder}

Gene expression is naturally organized by biological pathways: coordinated gene sets executing specific cellular processes. The Reactome database~\cite{reactome} captures this organization as a directed acyclic hierarchy from high-level cellular functions to specific molecular interactions. Encoding expression within this
structure grounds predictions in established biology, mitigates gene measurement noise through within-pathway aggregation, and reduces dimensionality to a scale suitable for small clinical cohorts. 

We curate a vocabulary from Reactome by filtering hierarchy depth, excluding categories unlikely to carry prognostic signal, enforcing pathway size constraints, and supplementing with MSigDB Hallmark gene sets~\cite{hallmark} to cover processes under represented in the filtered set. Redundant pathways with high gene overlap are removed via Jaccard similarity thresholding, yielding 662 pathways covering 4574 genes. Full details on our fully automated and reproducible curation pipeline are in SM~\ref{supp:pathways}.

\par\vspace{3pt}

\noindent\textbf{Graph Construction.} The curated gene--pathway membership structure is represented as a bipartite graph $\mathcal{G} = (\mathcal{V}_G \cup \mathcal{V}_P, \mathcal{E}_{GP})$ with gene nodes $\mathcal{V}_G$ and pathway nodes $\mathcal{V}_P$ connected by bidirectional membership edges $\mathcal{E}_{GP}$. Gene nodes are initialized with their scalar expression values $\mathbf{x}_g \in \mathbb{R}^1$; pathway nodes begin as zero vectors, ensuring their representations emerge entirely from constituent gene aggregation.

\par\vspace{3pt}

\noindent\textbf{Bipartite Message Passing.} Information propagates through $L{-}2$ GraphSAGE~\cite{Hamilton2017} layers with mean aggregation, each followed by LeakyReLU activation and dropout. The first layer lifts scalar gene features to the shared embedding dimension $d$; subsequent layers operate in $\mathbb{R}^d$:
\begin{equation}
    \mathbf{x}^{(l+1)} = \operatorname{GraphSAGE}\bigl(\mathbf{x}^{(l)},\,
    \mathcal{E}_{GP}\bigr), \quad l = 0, \ldots, L{-}2
\end{equation}
We use non-attentive mean aggregation in these layers: coherent expression shifts across functionally related genes provide more robust signal than individual gene magnitudes, and equal-weight aggregation captures this collective activity without introducing learned attention over insufficiently contextualized inputs. Because edges are bidirectional, each round simultaneously updates gene and pathway representations, and a gene participating in multiple pathways implicitly carries information between them, providing cross-pathway communication impossible under independent per-pathway encoding.

\par\vspace{3pt}

\noindent\textbf{Interpretable Gene--Pathway Attention.} After layer normalization, a final GATv2~\cite{GAT,GATv2} attention layer computes gene importance weights:
\begin{equation}
    \mathbf{x}^{(L)} = \operatorname{GATv2}\bigl(
    \operatorname{LayerNorm}(\mathbf{x}^{(L{-}1)}),\,
    \mathcal{E}_{GP}\bigr)
    \label{eq:gat}
\end{equation}
Reserving learned attention for this final layer ensures the GATv2 mechanism operates on enriched GraphSAGE representations rather than raw scalar values, enabling it to assess gene relevance in the context of overall pathway activity rather than from expression magnitude alone. The dynamic coefficients $\alpha_{g \to p}$ provide per-pathway gene-level interpretability directly from the architecture. Pathway embeddings $\{\mathbf{z}_p\}_{p=1}^{P} \in \mathbb{R}^{P \times d}$ are extracted from this final layer.

\par\vspace{3pt}

\noindent\textbf{Pathway Importance Gating.} A learnable gating network $\phi: \mathbb{R}^d \to \mathbb{R}$ maps each pathway embedding to a scalar score, and softmax normalization across pathways yields importance weights $w_p$. The patient-level genomic embedding is the importance-weighted aggregation $\bar{\mathbf{z}}_{\mathrm{pathway}}$:
\begin{equation}
    w_p = \operatorname{softmax}_p\bigl(\phi(\mathbf{z}_p)\bigr), \qquad \bar{\mathbf{z}}_{\mathrm{pathway}} = \sum_{p=1}^{P} w_p\, \mathbf{z}_p
    \label{eq:pathway_gate}
\end{equation}
The gate weights $\{w_p\}_{p=1}^{P}$ provide a built-in measure of pathway importance for each patient, directly interpretable without post-hoc attribution. Both individual pathway embeddings $\{\mathbf{z}_p\}_{p=1}^{P}$ and the aggregated $\bar{\mathbf{z}}_{\mathrm{pathway}}$ are passed forward: the former participate in cross-modal attention (Section~\ref{sec:fusion}), the latter contributes to the final fused representation.

\subsection{Cross-Modal Fusion}
\label{sec:fusion}

Tissue morphology is a consequence of underlying molecular programs: gene expression drives cellular differentiation, proliferation, and microenvironment remodeling, which collectively manifest as observed histological patterns. Our fusion mechanism reflects this asymmetry through cross-attention in which morphological prototypes query pathway embeddings, modeling the directional relationship from molecular context to morphological manifestation.

\par\vspace{3pt}

\noindent\textbf{Cross-Modal Attention.} Prototype tokens query pathway embeddings via multi-head scaled dot-product attention:
\begin{equation}
    \mathbf{A} = \operatorname{softmax}\!\left(\frac{\mathbf{T}\mathbf{W}_Q (\mathbf{Z}\mathbf{W}_K)^\top}{\sqrt{d}}\right), \qquad \tilde{\mathbf{T}} = \mathbf{A}\,\mathbf{Z}\mathbf{W}_V
    \label{eq:attention_matrix}
\end{equation}
where $\mathbf{T} \in \mathbb{R}^{K \times d}$ are prototype tokens, $\mathbf{Z} \in \mathbb{R}^{P \times d}$ are pathway embeddings, and $\mathbf{W}_Q, \mathbf{W}_K, \mathbf{W}_V \in \mathbb{R}^{d \times d}$ are learned projections. The attention matrix $\mathbf{A} \in \mathbb{R}^{K \times P}$ is the central interpretable output: each entry $a_{kp}$ quantifies the association between image prototype $k$ and biological pathway $p$ for a given patient.

\par\vspace{3pt}

\noindent\textbf{Post-Attention Gating and Combination.} A learned gating network computes importance weights over the attended prototype representations:
\begin{equation}
    w^{\mathrm{f}}_k = \operatorname{softmax}_k\bigl(\psi_f(\tilde{\mathbf{t}}_k)\bigr), \qquad \bar{\mathbf{z}}_{\mathrm{cross}} = \sum_{k=1}^{K} w^{\mathrm{f}}_k\, \tilde{\mathbf{t}}_k
    \label{eq:fusion_gates}
\end{equation}
These fusion gate weights $w^{\mathrm{f}}_k$ are distinct from the WSI gate weights $w_k$: the latter assess morphological importance in isolation, the former after pathway context has been incorporated. Comparing the two shows how cross-modal context reshapes the model's morphological priorities.

The final patient-level representation concatenates three layer-normalized (LN) streams:
\begin{equation}
    \mathbf{z}_{\mathrm{fused}} = \operatorname{MLP}\!\Big(\big[\operatorname{LN}(\bar{\mathbf{z}}_{\mathrm{pathway}})\;\|\;\operatorname{LN}(\bar{\mathbf{z}}_{\mathrm{cross}})\;\|\;\operatorname{LN}(\bar{\mathbf{z}}_{\mathrm{wsi}})\big]\Big)
\end{equation}
where $\|$ denotes concatenation and the MLP is a bottleneck projection with ReLU and dropout. The three streams carry complementary signals: $\bar{\mathbf{z}}_{\mathrm{pathway}}$ captures molecular information that may not manifest morphologically, $\bar{\mathbf{z}}_{\mathrm{wsi}}$ captures morphological patterns that may not be pathway-driven, and $\bar{\mathbf{z}}_{\mathrm{cross}}$ captures their associations. A linear classifier maps the fused representation to discrete survival bin logits.

\subsection{Interpretability Signals}
\label{sec:interpretability}

Because every prediction flows through semantically grounded representations, the architecture produces a complete attribution chain linking molecular programs to tissue morphology without post-hoc methods. Table~\ref{tab:signals} summarizes the interpretability signals emerging directly from inference; we describe how they compose below.

\begin{table}[t]
\centering
\caption{Interpretability signals produced during standard inference.}
\label{tab:signals}
\small
\resizebox{0.7\linewidth}{!}{%
\begin{tabular}{@{}clll@{}}
\toprule
 & \textbf{Signal} & \textbf{Source} & \textbf{Interpretation} \\
\midrule
\multicolumn{4}{@{}l}{\textit{Morphological}} \\[2pt]
\textsc{i} & $\alpha_{nk}$ & Soft assignment & Patch-to-prototype mapping \\
\textsc{ii} & $w_k$ & Prototype gate & Morphological importance (pre-fusion) \\
\midrule
\multicolumn{4}{@{}l}{\textit{Genomic}} \\[2pt]
\textsc{iii} & $\alpha_{g \to p}$ & GATv2 final layer & Gene importance per pathway \\
\textsc{iv} & $w_p$ & Pathway gate & Pathway prognostic relevance \\
\textsc{v} & $f(\alpha_{g \to p},\, w_p)$ & GATv2 derived & Overall gene importance \\
\midrule
\multicolumn{4}{@{}l}{\textit{Cross-modal}} \\[2pt]
\textsc{vi} & $\mathbf{A} \in \mathbb{R}^{K \times P}$ & Cross-attention & Prototype--pathway associations \\
\textsc{vii} & $\mathbf{A}_{k,:}$ & Rows of $\mathbf{A}$ & Per-prototype pathway ranking \\
\textsc{viii} & $w^{\mathrm{f}}_k$ & Fusion gate & Morphological importance (post-fusion) \\
\bottomrule
\end{tabular}}
\end{table}

\par\vspace{3pt}

\noindent\textbf{Morphological attribution.} Prototype gate weights $w_k$ assess morphological importance from tissue patterns alone, while fusion gate weights $w^{\mathrm{f}}_k$ assess importance after pathway context is incorporated. Comparing the two shows whether a prototype's prognostic relevance is intrinsically morphological or emerges through association with specific molecular programs.

\par\vspace{3pt}

\noindent\textbf{Genomic attribution.} GATv2 attention coefficients $\alpha_{g \to p}$ quantify each gene's importance within each of its pathways, while pathway gate weights $w_p$ identify which biological programs the model considers most prognostically relevant. Together they yield complete genomic attribution from individual genes through to pathway-level programs.

\par\vspace{3pt}

\noindent\textbf{Spatial overlays.} Patch-prototype assignments combined with the cross-modal attention matrix enable spatial overlays at multiple levels of granularity: prototype segmentation maps, pathway-colored tissue maps, and continuous single-pathway or single-gene heatmaps. All derive from inference-time signals; construction details are in SM~\ref{supp:interpretability}.

\par\vspace{3pt}

\noindent\textbf{Population-level analysis.} To identify signals systematically associated with outcomes, we employ rank-based statistical testing with appropriate handling of cross-validation structure. For pathways and genes, whose identities are consistent across folds, Mann-Whitney U tests are conducted within each fold and combined via Stouffer's weighted $Z$ method with FDR correction. For prototypes rank analysis is performed at the individual fold level. Full details are in SM~\ref{supp:statistics}.

\subsection{Training and Evaluation}
\label{sec:training}

\noindent\textbf{Survival objective.} We adopt the discrete-time survival framework of Zadeh and Schmidt~\cite{zadeh}, partitioning continuous survival times into $B=4$ quantile-based intervals. The model outputs logits $\mathbf{h} \in \mathbb{R}^B$, from which the discrete survival function is:
\begin{equation}
    S(b) = \prod_{j=1}^{b} \bigl(1 - \sigma(h_j)\bigr)
\end{equation}
Training minimizes negative log-likelihood over observed and censored patients. Patient-level risk scores are computed as $r = -\sum_{b=1}^{B} S(b)$, where the negation ensures higher scores correspond to worse prognosis.

\par\vspace{3pt}

\noindent\textbf{Evaluation protocol.} We follow the 5-fold cross-validation splits from SurvPath~\cite{survpath}, with model selection by validation C-index per fold. Patch features are pre-extracted using UNI-2h~\cite{UNI} at $D = 1536$ dimensions. All baselines were reimplemented within a unified codebase sharing identical preprocessing, survival discretization, pathway vocabulary, and evaluation procedure, ensuring comparisons reflect purely architectural differences. Hyperparameters in SM~\ref{supp:hyperparams}.

\par\vspace{3pt}

\noindent\textbf{Datasets and baselines.} We evaluate on five TCGA cohorts (BRCA, BLCA, COADREAD, HNSC, STAD; statistics in SM~\ref{supp:datasets}). Multimodal baselines include PORPOISE~\cite{porpoise}, MCAT~\cite{mcat}, MOTCAT~\cite{motcat}, SurvPath~\cite{survpath}, PIBD~\cite{pibd}, and MMP~\cite{mmp}; unimodal baselines and ProtoPathway's own unimodal ablations are reported in Table~\ref{tab:tcga_cindex}. To validate the fusion design, we compare cross-attention against concatenation, bilinear, and gated alternatives within the same architecture in SM~\ref{supp:ablation}. We showcase ProtoPathway's computational efficiency in SM~\ref{supp:efficiency}. 


\section{Experimental Results}

\subsection{Quantitative Performance}
\label{sec:quant}

\begin{figure}[h]
    \centering
    \includegraphics[width=1\linewidth]{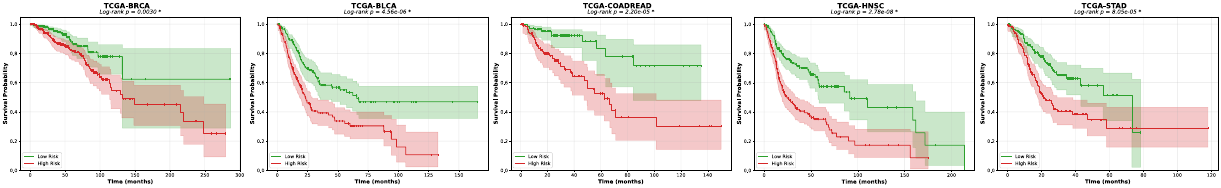}
    \caption{Kaplan-Meier survival curves for ProtoPathway across five TCGA cohorts. Patients stratified into high-risk (red) and low-risk (green) groups by median predicted risk score. All separations significant at $p<0.005$ (log-rank test).}
    \label{fig:km_curves}
\end{figure}

Table~\ref{tab:tcga_cindex} reports C-index results across all cohorts. ProtoPathway achieves the highest overall C-index (0.670), outperforming MCAT (0.662) and MMP (0.659), and ranks first on four of five cohorts; COADREAD is the sole exception, where SurvPath leads (0.755 vs.\ 0.740). On HNSC, ProtoPathway (0.642) is the only multimodal method to clearly surpass the gene-only SNN baseline (0.635), suggesting effective cross-modal integration on a cohort where prior multimodal approaches have struggled. Kaplan-Meier analysis (Fig.~\ref{fig:km_curves}) confirms significant risk stratification across all five cohorts ($p < 0.005$), with BLCA and HNSC showing particularly clean separation.

\begin{table}[h]
  \centering
  \caption[ProtoPathway performance on TCGA multimodal survival prediction benchmarks]{\textbf{ProtoPathway performance on TCGA multimodal survival prediction benchmarks.} Concordance index (C-index) results across five TCGA cancer cohorts comparing ProtoPathway against established multimodal survival methods. Standard error shown across cross-validation folds. Best score in \textbf{bold}, second best \underline{underlined}.}
  \label{tab:tcga_cindex}
  \setlength{\tabcolsep}{3pt}
  \renewcommand\arraystretch{1.12}
  \small
  \resizebox{\linewidth}{!}{%
  \begin{tabular}{r c c c c c c}
    \toprule
    & \textbf{BRCA} $\uparrow$ & \textbf{BLCA} $\uparrow$ & \textbf{COADREAD} $\uparrow$ & \textbf{HNSC} $\uparrow$ & \textbf{STAD} $\uparrow$ & \textbf{Overall} $\uparrow$ \\
    & \scriptsize(N=714) & \scriptsize(N=359) & \scriptsize(N=227) & \scriptsize(N=392) & \scriptsize(N=318) & \\
    \midrule

    \multicolumn{7}{@{}l}{\textit{WSI Modality}}\\
    ABMIL \cite{abmil}
       & 0.602$\pm$0.075 & 0.593$\pm$0.029 & 0.691$\pm$0.172 & 0.544$\pm$0.057 & 0.608$\pm$0.018 & 0.608 \\
    TransMIL \cite{transmil}
      & 0.575$\pm$0.125 & 0.596$\pm$0.057 & 0.717$\pm$0.095 & 0.546$\pm$0.063 & 0.631$\pm$0.106 & 0.613 \\
    DSMIL \cite{dsmil}
      & 0.627$\pm$0.062 & 0.641$\pm$0.071 & 0.660$\pm$0.148 & 0.604$\pm$0.028 & 0.627$\pm$0.049 & 0.632 \\
    $\text{PROTOPATH}_{\text{wsi}}$
      & 0.582$\pm$0.052 & 0.633$\pm$0.062 & 0.702$\pm$0.140 & 0.541$\pm$0.068 & 0.614$\pm$0.059 & 0.614 \\
    \addlinespace[2pt]
    \midrule
    
    \multicolumn{7}{@{}l}{\textit{Gene Modality}}\\
    MLP
      & 0.606$\pm$0.093 & 0.638$\pm$0.067 & 0.656$\pm$0.157 & 0.623$\pm$0.057 & 0.626$\pm$0.079 & 0.630 \\
    SNN~\cite{snn}
      & 0.592$\pm$0.033 & 0.645$\pm$0.067 & 0.652$\pm$0.093 & \underline{0.635$\pm$0.026} & 0.623$\pm$0.036 & 0.629 \\
    $\text{PROTOPATH}_{\text{gene}}$
      & 0.582$\pm$0.052 & 0.604$\pm$0.031 & 0.629$\pm$0.115 & 0.557$\pm$0.088 & 0.614$\pm$0.078 & 0.597  \\
    \addlinespace[2pt]
    \midrule
    
    \multicolumn{7}{@{}l}{\textit{Multimodal}}\\
    PIBD \cite{pibd}
      & 0.621$\pm$0.066 & 0.571$\pm$0.039 & 0.680$\pm$0.149 & 0.541$\pm$0.049 & 0.612$\pm$0.090 & 0.610 \\
    MOTCat \cite{motcat}
      & 0.581$\pm$0.028 & 0.642$\pm$0.057 & 0.659$\pm$0.168 & 0.587$\pm$0.037 & 0.660$\pm$0.061 & 0.626 \\
    PORPOISE \cite{porpoise}
      & 0.621$\pm$0.072 & 0.642$\pm$0.036 & 0.702$\pm$0.113 & 0.617$\pm$0.057 & 0.669$\pm$0.046 & 0.650 \\
    MMP \cite{mmp}
      & 0.637$\pm$0.064 & \underline{0.645$\pm$0.045} & 0.732$\pm$0.085 & 0.610$\pm$0.026 & \underline{0.673$\pm$0.072} & 0.659 \\
    SURVPATH \cite{survpath}
      & 0.641$\pm$0.046 & 0.642$\pm$0.021 & \textbf{0.755$\pm$0.112} & 0.604$\pm$0.062 & 0.658$\pm$0.075 & 0.660 \\
    MCAT \cite{mcat}
      & \underline{0.646$\pm$0.076} & 0.643$\pm$0.052 & 0.726$\pm$0.113 & 0.621$\pm$0.040 & 0.672$\pm$0.058 & \underline{0.662} \\
    PROTOPATH (ours)
      & \textbf{0.649$\pm$0.050} & \textbf{0.646$\pm$0.075} & \underline{0.740$\pm$0.132} & \textbf{0.642$\pm$0.047} & \textbf{0.674$\pm$0.069} & \textbf{0.670} \\
    \bottomrule
  \end{tabular}
}
\end{table}

\par\vspace{3pt}

\noindent The unimodal ablations shows an instructive asymmetry: \textsc{ProtoPath}\textsubscript{gene} (0.597) falls below both MLP (0.630) and SNN (0.629), yet the full model recovers to the best overall score (0.670). The bipartite graph topology acts as a strong structural regularizer, constraining each pathway embedding to aggregate only its constituent genes via fixed Reactome edges. Unimodal training collapses these into a single pooled vector, discarding the very structure that distinguishes them; cross-modal attention, by contrast, queries each pathway embedding individually, turning those same constraints into an asset by providing decomposed, semantically localized tokens the fusion module can selectively attend to. A parallel pattern holds on the WSI side: \textsc{ProtoPath}\textsubscript{wsi} (0.614) outperforms ABMIL (0.608) and TransMIL (0.613) at a fraction of DSMIL's cost (0.632; 5$\times$ parameters, 7.5$\times$ FLOPs; \textit{cf}.\ Tab.~\ref{tab:efficiency}), again favoring structured representations over raw unimodal capacity. We further speculate that multimodal training amplifies this effect, as gradients flowing back through cross-attention implicitly encourage pathway representations that complement morphological features.

\par\vspace{3pt}

\noindent Overall, this performance comes at substantially lower computational cost: 480K parameters, 3.9G FLOPs, 325\,MB VRAM, and 13.6\,ms per patient, a 28--50$\times$ training speedup over MCAT (677K params, 5.6G FLOPs, 289\,MB, 380\,ms), SurvPath (474K params, 29.2G FLOPs, 818\,MB, 476\,ms), and MMP (427K params, 1{,}817\,MB, 432\,ms), stemming directly from prototype-based compression over $K \!\ll\! N$ tokens (see\ SM~\ref{supp:efficiency} for a detailed look at comparative efficiency).

\subsection{Biological Interpretability}
\label{sec:pathways}

We assess whether ProtoPathway's learned representations capture biologically meaningful structure, using BLCA as a case study where the tumor immune microenvironment is well-characterized and clinically actionable.

\par\vspace{3pt}

\noindent\textbf{Morphological prototypes.} The $K\!=\!16$ learned prototypes decompose the WSI into interpretable tissue categories without tissue-type supervision. The top exemplar patches per prototype reveal coherent morphological groupings (see SM~\ref{supp:supp_proto} and Fig.~\ref{fig:proto_exemplars} for descriptions of the prototypes, with exemplar patches): tumor and tumor-associated stroma (Proto~0, 8, 9, 10, 12), muscle and connective tissue (Proto~5, 11), adipose tissue (Proto~7, 13), and immune-infiltrated or necrotic regions (Proto~2, 3, 6). In bladder cancer, these distinctions are clinically significant: invasion depth into the muscularis propria defines the boundary between non-muscle-invasive and muscle-invasive disease~\cite{babjuk2022european}.

\par\vspace{3pt}

\begin{figure}[t]
  \centering
  \includegraphics[width=\linewidth]{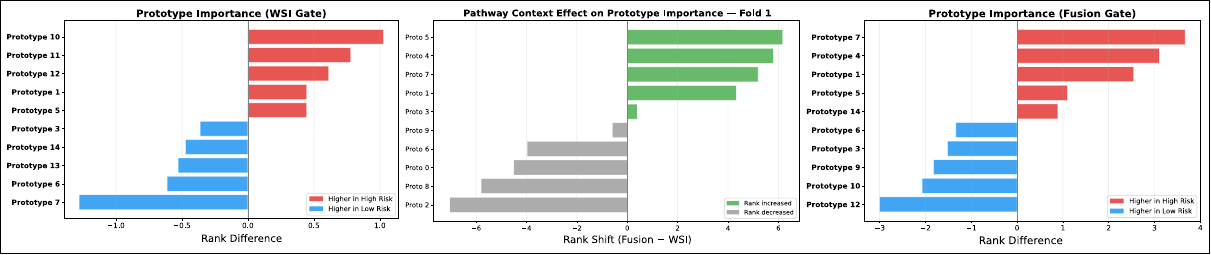}
  \caption{\textbf{Prototype analysis in BLCA.} Gating shift analysis reveals which morphological prototypes the model attends to before (WSI gate, left) and after (fusion gate, right) molecular pathway context is incorporated, with the rank shift between the two (center). Muscle and connective tissue prototypes (Proto~5, 11) gain importance after fusion (green), while tumor prototypes (Proto~2, 8) lose rank (grey), indicating that pathway context shifts attention toward the tumor microenvironment and away from morphologically prominent tumor regions. This aligns with established BLCA prognostics, where muscle invasion defines staging (T1 vs T2+) and stromal-immune composition carries independent prognostic value beyond tumor morphology.}
  \label{fig:proto_importance}
\end{figure}

\noindent\textbf{Gating shift.} Comparing prototype importance before (WSI gate $w_k$) and after (fusion gate $w^{\mathrm{f}}_k$) cross-modal attention reveals how pathway context reshapes morphological priorities (Fig.~\ref{fig:proto_importance}). Connective, adipose and smooth muscle prototypes gain importance after fusion, while tumor and immune-infiltrated prototypes lose rank. This suggests that once molecular context is accounted for, the structural microenvironment surrounding the tumor, particularly the muscularis propria invasion front and perivesical connective tissue, carries more survival-relevant information than morphologically dominant tumor and immune regions alone. The three-gate design makes this reweighting explicit and quantifiable.

\par\vspace{3pt}

\begin{figure}[htb]
    \centering
    \includegraphics[width=1\linewidth]{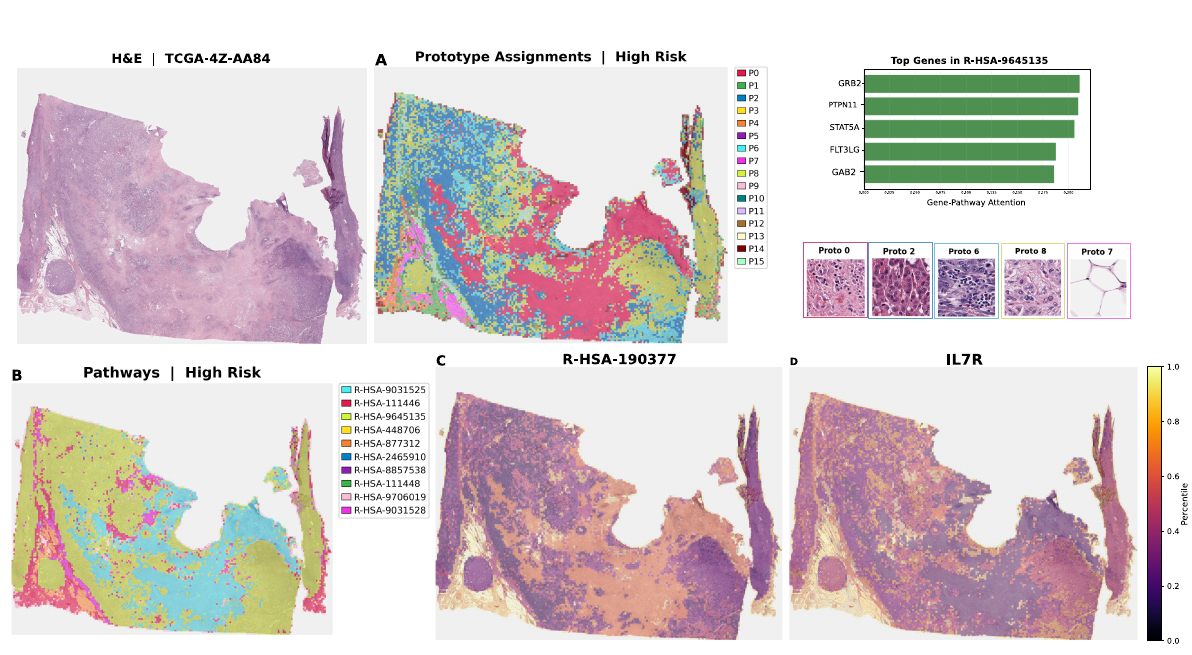}
    \caption{\textbf{Spatial attribution for a high-risk BLCA case (TCGA-4Z-AA84).} \textbf{Top row:} (left) H\&E image, (center) prototype segmentation showing spatially coherent tissue compartments, and (right) gene-level drilldown for the dominant pathway at the immune-rich periphery (STAT5 activation)). \textbf{Bottom row:} (left) pathway overlay assigning each prototype its top risk-relevant pathway, and (center) rank-transformed single-pathway (FGFR2b), (right) and single-gene (IL7R) heatmaps.}
    \label{fig:spatial}
\end{figure}

\noindent\textbf{Spatial heatmaps.} ProtoPathway provides attribution across the full biological hierarchy, from genes through pathways to spatial tissue locations. We illustrate this on our model determined highest-risk BLCA case (TCGA-4Z-AA84, Fig.~\ref{fig:spatial}), with prototype, pathway and gene overlays.


\par\vspace{3pt}

 \noindent\textbf{Prototype overlay.} Each patch is assigned to its nearest prototype, producing a spatial map of tissue composition (Fig.~\ref{fig:spatial}A). The prototypes partition the slide into histologically coherent regions: P0 (red/pink) covers the bulk invasive tumor and associated desmoplasia/necrosis, P2 (blue) captures dense lymphoid infiltrate and immune-rich margin in the upper-left, and P7 (magenta) isolates perivesical adipose tissue in the lower-left. P6 (cyan) and P8 (yellow-green) mark transitional zones at compartment boundaries and tissue edges, while the remaining prototypes capture smaller peripheral and artefactual regions.

 \par\vspace{3pt}
 
\noindent\textbf{Pathway overlay.} Each prototype inherits its top, population level (\textit{cf.} SM~\ref{supp:statistics}, risk-relevant pathway from the cross-modal attention, projecting molecular program identity onto the tissue map (Fig.~\ref{fig:spatial}B). Different compartments receive distinct pathway associations: the tumor bulk maps to cholesterol metabolism (NR1H2/NR1H3 regulation of cholesterol uptake), the immune-rich periphery to cytokine signaling (STAT5 activation, with gene drilldown confirming core JAK-STAT components: GRB2, PTPN11, STAT5A, FLT3LG), and sparse necrotic foci to interleukin-1 processing via the NLRP3 inflammasome. At the tissue margins, a second NR1H2/NR1H3 program appears, this time linked to triglyceride lipolysis in adipose tissue, consistent with the perivesical fat expected at the boundary of a bladder resection specimen.

\par\vspace{3pt}

\noindent\textbf{Single-pathway and single-gene overlays.} To move beyond discrete prototype-level assignments, we visualize continuous, rank-transformed attention for a single pathway (FGFR2b ligand binding; Fig.~\ref{fig:spatial}C) and a single gene (IL7R; Fig.~\ref{fig:spatial}D). These two overlays exhibit complementary spatial patterns. FGFR2b attention concentrates in the tumor bulk and is low at the immune-rich periphery; IL7R attention shows the inverse, highest at the periphery and lowest in the tumor core. This complementarity is biologically grounded. FGFR2b acts as a tumor suppressor in urothelial carcinoma, where its loss marks aggressive disease~\cite{diez1997}: the model identifies the tumor compartment as the region where this pathway's status is most prognostically relevant. IL7R maintains T-cell survival and homeostasis: its attention localizes to the compartment where immune cells concentrate. The model thus resolves two distinct prognostic programs within a single slide: tumor-intrinsic growth factor signaling and microenvironment immune activity.

\par\vspace{3pt}


\begin{figure}[htb]
    \centering
    \includegraphics[width=1\linewidth]{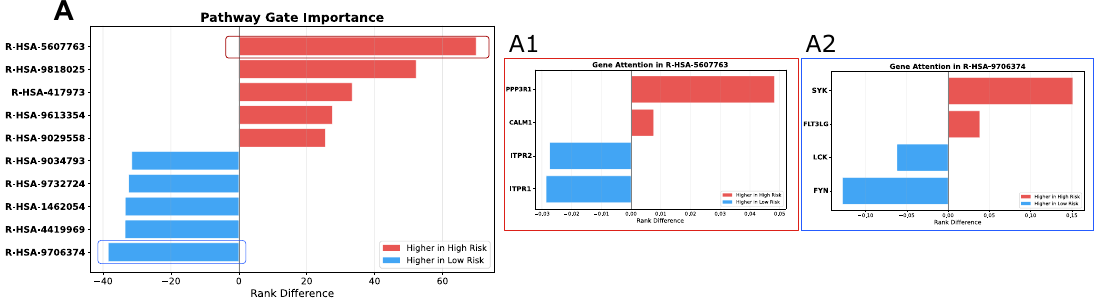}
    \caption{\textbf{Pathway and gene-level attribution in BLCA.} \textbf{(A)}~Pathway gate importance: rank difference between high-risk (red) and low-risk (blue) groups. High risk pathways - R-HSA-5607763: CLEC7A/NFAT activation, R-HSA-9818025: NFE2L2/NRF2, R-HSA-417973: adenosine signaling, R-HSA-9613354/R-HSA-9029558: lipid metabolism reprogramming. Low risk pathways - R-HSA-9706374: FLT3/SRC signalling, R-HSA-9732724: interferon-$\gamma$ signaling,  R-HSA-4419969: apoptotic program engagement. \textbf{(A1,\,A2)}~Gene-level analysis within the top low-risk pathway (A1: FLT3/SRC signaling) and top high-risk pathway (A2: CLEC7A/NFAT activation).}
    \label{fig:pathway_genes}
\end{figure}


\noindent\textbf{Pathway-level analysis.} Using fold-stratified population analysis, we compare pathway attention distributions between model-predicted high-risk and low-risk groups. Fig.~\ref{fig:pathway_genes}A shows the top-ranked pathways by risk direction. The model assigns highest risk to pathways involved in innate immune signalling (CLEC7A/NFAT activation~\cite{Kawahara2015}), oxidative stress response (NFE2L2 /NRF2~\cite{hayden2014nrf2}), immunosuppressive adenosine signalling~\cite{Vijayan2017_adenosine}, and lipid metabolism reprogramming \cite{Zhu2021_lipid}. These are established markers of aggressive, treatment-resistant bladder cancer. Conversely, the top low-risk pathways center on adaptive immune activation: FLT3 signaling (whose ligand FLT3LG promotes CD8+ T-cell activation in BLCA \cite{zhang2024bcg}, interferon-$\gamma$ signaling \cite{gillezeau2022interferon}, and apoptotic programme engagement, consistent with the role of apoptosis in cancer progression~\cite{hanahan2011hallmarks}. Without any explicit immune labels, the model recovers a known prognostic distinction: tumours dominated by immunosuppressive innate signalling carry worse prognosis than those with active adaptive immunity~\cite{kamoun2020consensus,wu2020profiles}.

\par\vspace{3pt}

\noindent\textbf{Pathway genes drill-down.} The bipartite GNN's gene-to-pathway attention enables inspection of individual genes within each pathway (Fig.~\ref{fig:pathway_genes}B). Within the top high-risk pathway (CLEC7A/NFAT), the model assigns highest importance to PPP3R1 and CALM1: the regulatory subunit and upstream activator of calcineurin through which the entire cascade converges~\cite{hogan2003}. Within the top low-risk pathway (FLT3/SRC), the highest-ranked genes are LCK and FYN, the proximal T-cell receptor kinases whose expression marks cytotoxic T-cell infiltration associated with favorable prognosis in BLCA~\cite{zheng2021lck}. In both cases, attention concentrates on mechanistic bottlenecks rather than peripheral pathway members.

\par\vspace{3pt}

\noindent\textbf{Limitations} The bipartite encoder's strong structural priors introduce additional optimization complexity relative to unconstrained encoders. This is an inherent trade-off: the same constraints that complicate optimization are what yield the framework's efficiency, semantically grounded tokens, and gene-level importance. Beyond architectural considerations, the input modalities themselves impose interpretive limits. The framework operates on bulk transcriptomic profiles, which average over cell populations. The spatial overlays linking pathways to tissue regions should therefore be interpreted as inferred cross-modal associations projected onto prototype locations, not as direct measurements of localized gene expression, and spatial transcriptomic data would be needed for direct validation. A related caveat concerns the interpretation of attention itself. The framework identifies which genes and pathways the model attends to within risk groups, but attention magnitude does not encode the direction of expression change. Resolving this requires separate differential expression analysis.

\section{Conclusion
}\label{sec:conclusion}

ProtoPathway provides a complete interpretability chain from individual genes, through biological pathways, to specific tissue regions, while matching or exceeding existing methods on survival prediction across five cancer cohorts. This is enabled by bipartite message passing over gene-pathway relationships, learnable prototype-based morphological compression, and asymmetric cross-modal fusion in which prototypes query pathways, reflecting the biological direction from molecular programs to tissue morphology. The learned representations recover known prognostic signatures without explicit supervision and resolve distinct molecular programs across tissue compartments: cross-modal fusion redirects attention from morphologically dominant tumor and immune regions toward staging-relevant microenvironment structures, while continuous pathway and gene overlays reveal complementary prognostic programs localized to their expected tissue compartments. Spatial transcriptomics offers a natural extension: spatially resolved molecular profiles could validate the cross-modal associations that ProtoPathway currently learns without spatial supervision, and enable finer-grained fusion at the level of tissue neighborhoods rather than whole-slide summaries.

\section*{Acknowledgments}

We wish to thank Dr. Omnia Alwazzan for the many discussions on multimodal fusion models, which paved the way for this work. During this work A.G.S. received funding from the Wellcome Trust [218584/Z/19/Z]. This work acknowledges the support of the National Institute for Health and Care Research Barts Biomedical Research Centre (NIHR203330), a delivery partnership of Barts Health NHS Trust, Queen Mary University of London, St George’s University Hospitals NHS Foundation Trust and St George’s University of London. This work was also supported by the Engineering and Physical Sciences Research Council [grant number EP/Y009800/1], through funding from Responsible Ai UK (KP0016). This paper utilized Queen Mary’s Andrena HPC facility \cite{king_apocrita_2017}. The results shown here are in whole or part based upon data generated by the TCGA Research Network: \href{https://www.cancer.gov/tcga}{\texttt{https://www.cancer.gov/tcga}}.



%
%
\bibliographystyle{splncs04}
\bibliography{references}


\newpage

\clearpage
\setcounter{section}{0}
\renewcommand{\thesection}{\Alph{section}}
\setcounter{figure}{0}
\renewcommand{\thefigure}{S\arabic{figure}}
\setcounter{table}{0}
\renewcommand{\thetable}{S\arabic{table}}
\setcounter{equation}{0}
\renewcommand{\theequation}{S\arabic{equation}}

\setcounter{page}{1}

\section*{Supplementary Material}

This supplement provides full details on all components of ProtoPathway. SM~\ref{supp:kmeans} describes the $k$-means prototype initialization procedure. SM~\ref{supp:pathways} details the two-stage pathway and gene curation pipeline. SM~\ref{supp:interpretability} defines the eight interpretability signals extracted during inference and explains how they compose into spatial overlays. SM~\ref{supp:statistics} presents the rank-based statistical framework used for population-level analysis, including fold-stratified testing and Stouffer's meta-analytic combination. SM~\ref{supp:hyperparams} lists all hyperparameters. SM~\ref{supp:datasets} describes the five TCGA cohorts. SM~\ref{supp:ablation} reports ablation studies on modality branches, fusion mechanism, and prototype count. SM~\ref{supp:efficiency} compares computational cost across all methods. SM~\ref{supp:supp_proto} provides per-prototype morphological descriptions with exemplar patches.

\section{Prototype Initialization}
\label{supp:kmeans}

The learnable prototypes $\{\mathbf{c}_k\}_{k=1}^{K}$ described in Section~\ref{sec:wsi_encoder} are initialized from k-means centroids computed on training data. Because prototype assignment (Eq.~1) operates through cosine similarity in the projected space, initialization via cosine-distance clustering in the input space $\mathbb{R}^D$ provides a starting decomposition aligned with the similarity metric the model will refine during training.

\subsubsection{Training data sampling.}
For each cross-validation fold, up to $10^5$ patch embeddings are sampled without replacement from training slides only, proportional to each slide's patch count to ensure broad representation while preventing validation data leakage. Proportional counts are computed as $n_i = \max(1, \lfloor N_i \cdot 10^5 / N_{\mathrm{total}} \rfloor)$, where $N_i$ is the patch count of slide $i$ and $N_{\mathrm{total}}$ the total across all training slides. The $\max(1, \cdot)$ floor guarantees every training slide contributes at least one patch to the clustering sample.

\subsubsection{$\ell_2$ normalization and clustering.}
All sampled embeddings are $\ell_2$-normalized to unit norm ($\|\mathbf{h}\|_2 = 1$) before clustering. For unit vectors $\mathbf{u}$ and $\mathbf{v}$, squared Euclidean distance reduces to cosine distance:
\begin{equation}
    \|\mathbf{u} - \mathbf{v}\|^2 = 2\bigl(1 - \langle \mathbf{u}, \mathbf{v} \rangle\bigr)
    \label{eq:cosine_euclidean}
\end{equation}
This equivalence allows standard k-means, which minimizes squared Euclidean distance, to operate as cosine-distance clustering without algorithmic modification. Cosine distance is more appropriate than Euclidean distance for high-dimensional pathology features, where angular relationships encode morphological similarity while magnitude can vary with staining intensity or tissue density. Clustering uses \texttt{MiniBatchKMeans} with batch size 1024 and 10 \texttt{k-means++} restarts.

\subsubsection{Centroid post-processing.}
Converged centroids are re-normalized to unit norm. During k-means, centroids are computed as arithmetic means of their assigned vectors, which produces vectors with magnitude less than one because averaging partially cancels components that point in different directions. Re-normalization restores the unit-length constraint so that subsequent cosine comparisons depend purely on angular differences. The $K$ centroids are stored as 32-bit floating-point tensors and cached per fold for reproducibility across training runs, ensuring that restarted or resumed experiments use identical initialization.

\subsubsection{Initialization in the model.}
The cached centroids are passed to the PrototypeMIL module as $\texttt{init\_centroids} \in \mathbb{R}^{K \times D}$, where they initialize the learnable prototype parameter $\{\mathbf{c}_k\}_{k=1}^{K}$. From this point, the prototypes are trainable parameters updated end-to-end by the survival objective, free to diverge from their initial positions to capture prognostically relevant morphological variation that purely unsupervised clustering may not prioritize.

\subsubsection{Cross-fold prototype identity.}
Because centroids are computed independently per fold from different training subsets, and because end-to-end optimization permits prototypes to diverge from their initialization, the learned prototypes are not directly comparable across folds. This fold specificity is not inherently undesirable: it allows the model to adapt its morphological vocabulary to the particular tissue distribution in each training split. However, it precludes population-level statistical analysis of prototype importance across folds, in contrast to pathways and genes whose identities are fixed by the Reactome vocabulary. Prototype-level interpretability analysis is therefore conducted within individual folds. Enforcing cross-fold consistency during training, for example through diversity regularization encouraging prototypes to occupy distinct and reproducible regions of the feature space, anchor-based penalties biasing prototypes toward a shared reference vocabulary, or limited concept supervision over tissue-type annotations, remains an open direction for future work.

\section{Pathway and Gene Curation}
\label{supp:pathways}

The biological pathway vocabulary used by ProtoPathway is constructed through a fully automated, reproducible two-stage pipeline. Stage~1 builds a dataset-agnostic base pathway set from Reactome and MSigDB Hallmark gene sets. Stage~2 adapts this base set to each TCGA cohort by intersecting with the available gene expression data and constructing the bipartite graph consumed by the model. All parameters are specified in YAML configuration files and encoded in output filenames for full traceability. The pipeline is invoked via the unified Command Line Interface (CLI): \texttt{python main.py preprocess reactome} (Stage~1) and \texttt{python main.py preprocess genes} (Stage~2).

\subsection{Stage 1: Base Pathway Construction}
\label{sm:curation:stage1}

\subsubsection{Reactome Pathway Loading.} Reactome pathways (Version~83; 2,769 \textit{Homo sapiens} pathways) \cite{reactome} are loaded from the standard GMT file, which encodes each pathway as a gene set. The Reactome hierarchy is reconstructed as a directed acyclic graph (DAG) from the official \texttt{ReactomePathwaysRelation.txt} file, filtering for human pathways (prefix \texttt{R-HSA}). Pathway depths are computed as shortest-path distances from the root nodes of the DAG. Each pathway is additionally mapped to its top-level ancestor category (e.g., ``Immune System'', ``Signal Transduction'') via graph traversal.

\subsubsection{Hierarchy-Based Depth Selection.} 
\begin{figure}[h]
    \centering
    \includegraphics[width=0.6\linewidth]{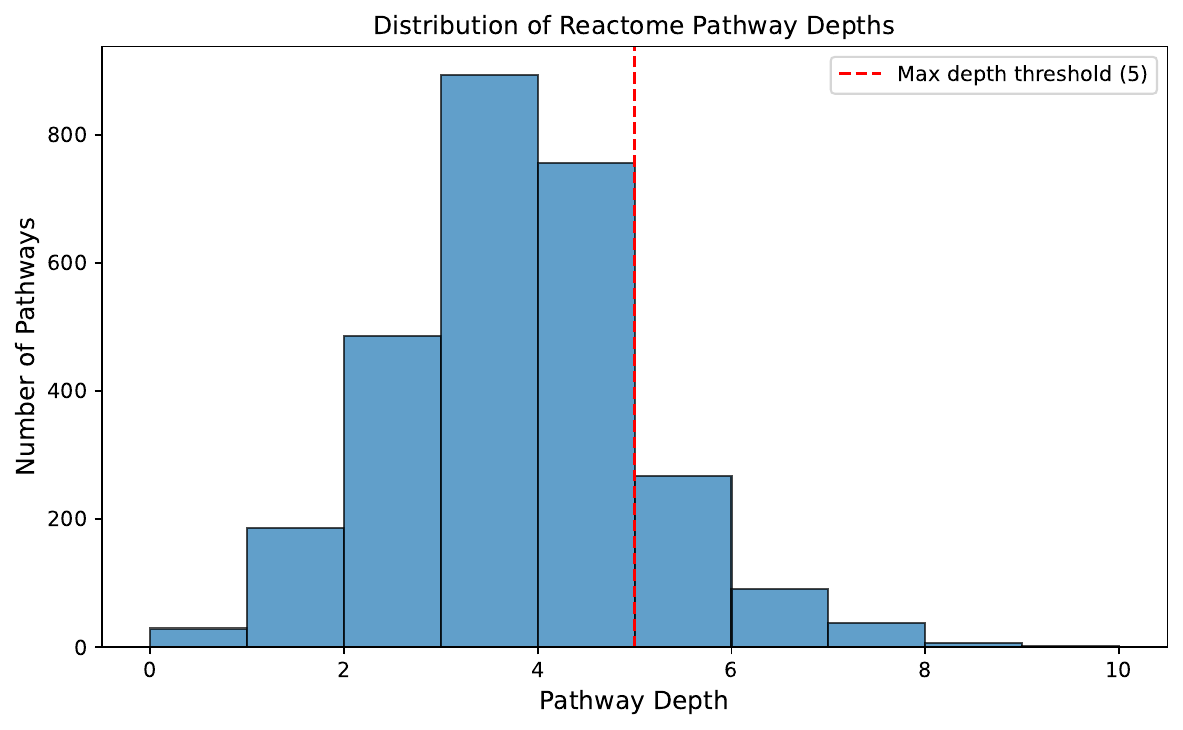}
        \caption[Reactome pathway hierarchy depth distribution and selection strategy]{\textbf{Reactome pathway hierarchy depth distribution and selection strategy.} The histogram displays the distribution of pathway depths within the Reactome hierarchical structure, with the vertical red line indicating the target depth of 5 selected for analysis. This depth represents an optimal balance between biological specificity and interpretable scope: shallower pathways (depths 1-3) encompass overly broad cellular processes that provide limited mechanistic insight, whilst deeper pathways (depths 6-8) represent highly specific molecular interactions that may not capture the systems-level perturbations observed of cancer.}
    \label{fig:pathway_depth}
\end{figure}

Pathways are selected at a target depth of~5 in the Reactome hierarchy, which balances biological specificity with sufficient gene membership for meaningful aggregation. For branches that do not extend to depth~5, the deepest available leaf node is retained, ensuring that shallower but terminally specific processes are not discarded. Pathways at depths greater than~5 are excluded as overly granular, whilst non-leaf pathways at depths less than~5 are excluded because their more specific children provide better resolution. We show the depth distribution of different Reactome branches in Fig.~\ref{fig:pathway_depth}.

\vspace{-20pt}

\subsubsection{Category Exclusion.} Entire top-level Reactome categories are removed when their constituent pathways are unlikely to carry prognostic signal for cancer survival, would introduce circularity, or represent housekeeping processes whose variation across patients is primarily technical rather than biological. The excluded categories are:

\begin{itemize}
    \item[\textbullet] \textbf{Pharmacology:} Drug ADME (drug absorption, distribution, metabolism, excretion; not endogenous biology).
    \item[\textbullet] \textbf{Disease-specific:} Disease (risk of circular reasoning, as these pathways are defined by the conditions we aim to predict, and include irrelevant non-cancer diseases).
    \item[\textbullet] \textbf{Housekeeping / basic cellular machinery:} Metabolism of proteins, Gene expression (Transcription), Organelle biogenesis and maintenance, Protein localization, Transport of small molecules, Vesicle-mediated transport, DNA Replication.
    \item[\textbullet] \textbf{Low cancer relevance:} Neuronal System, Sensory Perception, Muscle contraction, Digestion and absorption, Reproduction, Circadian clock.
    \item[\textbullet] \textbf{Broad mixed categories:} Metabolism, Developmental Biology (excluded in full, with targeted Hallmark gap-filling for cancer-relevant metabolic processes; see below).
\end{itemize}

\begin{figure}[h]
    \centering
    \includegraphics[width=1\linewidth]{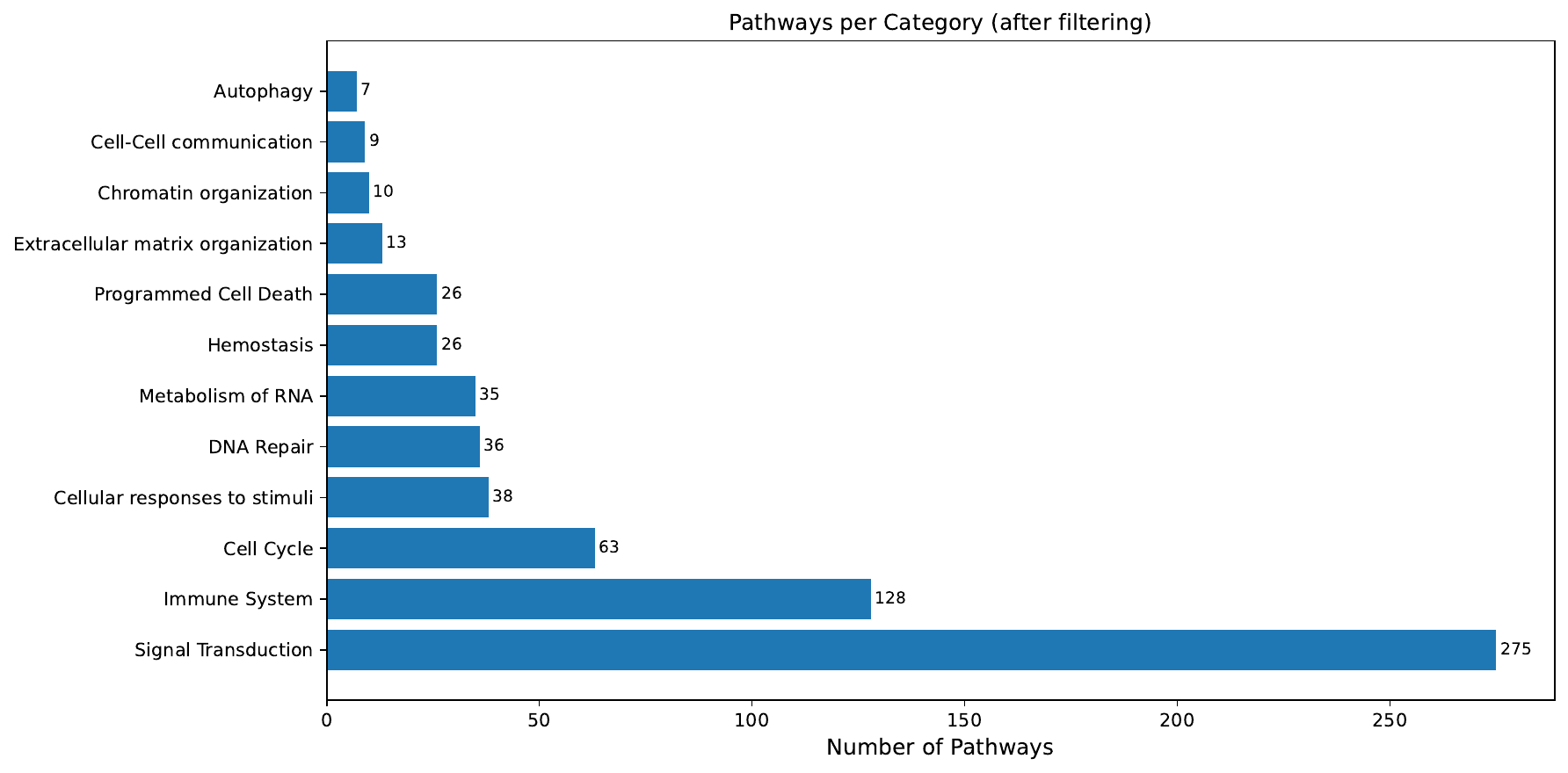}
    \caption{\textbf{Distribution of Reactome pathways across top-level categories after category exclusion.} The 666 retained pathways after category exclusion span 12 functional categories, dominated by Signal Transduction (275) and Immune System (128), with additional representation from Cell Cycle, Cellular responses to stimuli, DNA Repair, and Metabolism of RNA.}
    \label{fig:pathway_cat}
\end{figure}

This category-level filtering reduces the number of pathways from 2769 to 666 in a single principled step, substantially reducing the vocabulary before finer-grained size and redundancy filters are applied. In Figure~\ref{fig:pathway_cat}, we show the category distribution of the remaining pathways. 

\subsubsection{Size Filtering.} Pathways with fewer than 3 genes or more than 200 genes are excluded. The lower bound ensures sufficient context for message passing within the bipartite graph, as pathways with only 1--2 genes provide negligible aggregation benefit. The upper bound removes excessively broad pathway complexes that would dilute gene-specific signal during neighborhood aggregation. This step reduces the number of pathways to 634. We show the distribution of the number of genes per pathway in Figure~\ref{fig:pathway_dist}.

\begin{figure}[h]
    \centering
    \includegraphics[width=1\linewidth]{images/size_histogram_final.pdf}
    \caption{\textbf{Distribution of curated pathway sizes after filtering.} Histogram showing the number of genes per pathway for the 634 remaining Reactome pathways following hierarchy-based selection, category exclusion, size filtering (3-200 genes). The distribution exhibits a right-skewed pattern with most pathways containing fewer than 50 genes, while maintaining representation of larger pathway complexes up to the 200-gene threshold.}
    \label{fig:pathway_dist}
\end{figure}

\vspace{-20pt}

\subsubsection{Hallmark Supplementation.} Excluding entire Reactome categories risks removing cancer-relevant biological processes. Metabolism and Developmental Biology, for instance, contain pathways central to tumor progression (the Warburg effect, metabolic reprogramming, epithelial-mesenchymal transition) that would be lost under blanket category exclusion. To ensure comprehensive coverage of cancer-relevant biology, all 50 MSigDB Hallmark gene sets~\cite{hallmark} are integrated into the pathway vocabulary. The Hallmark collection provides broadly curated, non-redundant gene sets that span key oncogenic processes including proliferative signaling, inflammatory response, hypoxia, apoptosis, angiogenesis, and metabolic reprogramming, complementing the more granular Reactome pathways with process-level summaries. When a Hallmark gene set is integrated, any remaining Reactome pathway with Jaccard similarity $J = 1.0$ to that Hallmark set is removed to avoid exact duplication, though in practice the category exclusion step has already removed most overlapping Reactome pathways. This steps results in a pathway vocabulary of 684 pathway, with $\approx 93\%$ being Reactome and $7\%$ Hallamrk.

\subsubsection{Redundancy Removal.} Pairwise Jaccard similarity $J(A, B) = |A \cap B| / |A \cup B|$ is computed between all pathway gene sets. Pathway pairs with perfect overlap ($J = 1.0$) are consolidated. When selecting which pathway to retain from a redundant group, a hierarchy-aware priority ordering is applied: (1)~leaf nodes in the Reactome DAG are preferred over internal nodes, as they represent the most specific available characterisation; (2)~deeper pathways are preferred over shallower ones; (3)~larger gene sets are preferred; (4)~alphabetical ordering serves as a final tie-breaker. This ensures that the retained representative is maximally specific and informative.

\subsubsection{Base Pathway Statistics.} The Stage~1 pipeline produces a base set of curated pathways with associated gene memberships, saved with a parameter-encoded filename (e.g., \texttt{pathways\_base\_d5\_g3-200\_j100.pkl}) for reproducibility, as well as a pathway redundancy report. We show summary statistics of the curated pathways in Table~\ref{tab:pathway_stats}.

\begin{table}[h!]
    \centering
    \caption{\textbf{Base pathway vocabulary statistics (Stage~1).} Summary of the curated pathway set after hierarchy selection, category exclusion, size filtering, Hallmark supplementation, and redundancy removal. This base set is subsequently adapted per cohort in Stage~2, yielding the final 662 pathways reported in the main paper.}
    \label{tab:pathway_stats}
    \begin{tabular}{lr}
        \toprule
        \textbf{Statistic} & \textbf{Value} \\
        \midrule
        Total pathways & 684 \\
        \quad \quad Reactome & 634 \\
        \quad \quad Hallmark & 50 \\
        Total unique genes & 7{,}160 \\
        \midrule
        Mean genes per pathway & 35.7 \\
        Median genes per pathway & 19.0 \\
        \midrule
        Mean pathways per gene & 3.41 \\
        Genes in single pathway & 3{,}048 \\
        Genes in 10+ pathways & 409 \\
        \bottomrule
    \end{tabular}
\end{table}

\subsection{Stage 2: Per-Dataset Adaptation.} Stage~2 adapts the base pathway set to each TCGA cohort individually, accounting for differences in gene coverage across datasets.

\subsubsection{Gene Expression Data.} For all five TCGA cohorts, gene expression data is obtained directly from the SurvPath repository~\cite{survpath}, ensuring identical input data across all methods in our benchmark. The SurvPath expression files provide a common set of 4{,}999 genes across all cohorts. Because the SurvPath data has already been preprocessed and normalized, the optional expression filtering steps in our pipeline (CPM thresholding, log$_2$ transformation, mean-centering, variance selection) are bypassed, avoiding any additional preprocessing variation that could confound comparisons.

\subsubsection{Gene Coverage Filtering.} 
\begin{figure}[h]
    \centering
    \includegraphics[width=\linewidth]{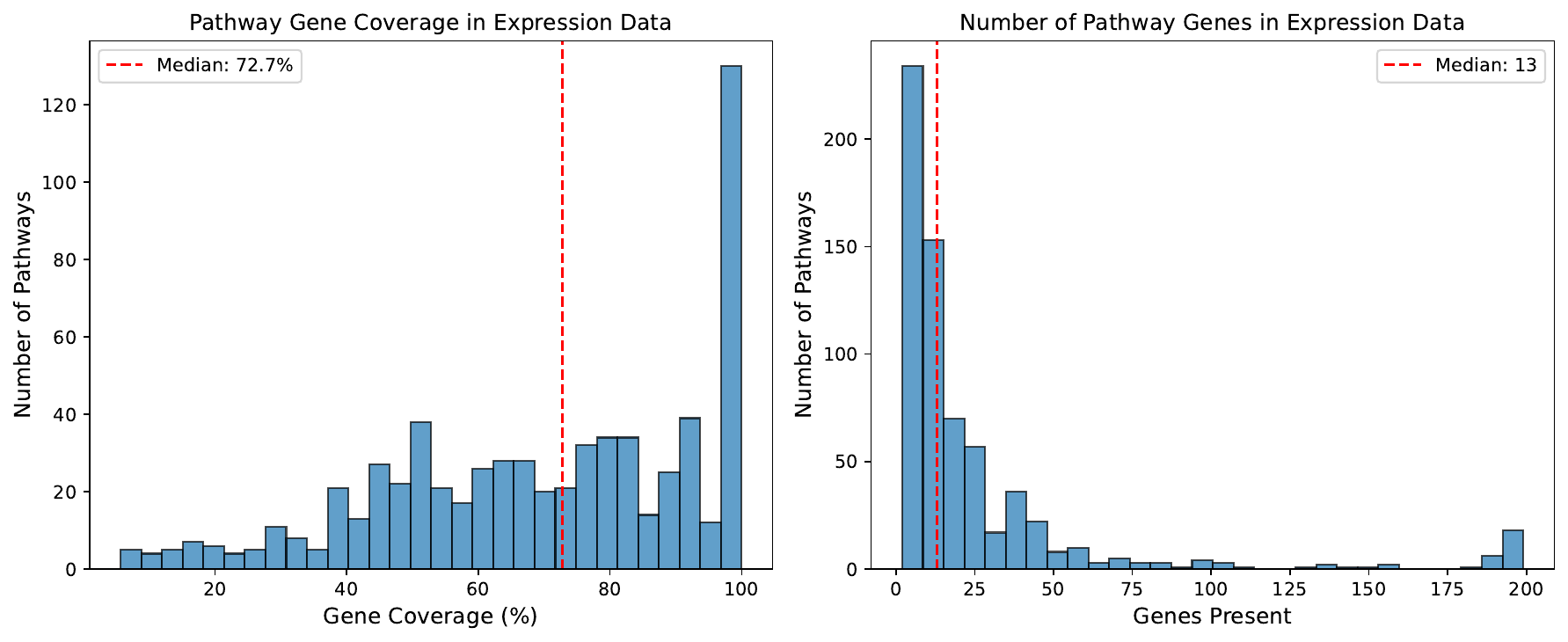}
    \caption{\textbf{Pathway gene coverage in the SurvPath expression data.} Left: percentage of each pathway's original gene set represented among the 4{,}999 SurvPath features (median 72.7\%). Right: absolute number of genes present per pathway (median 13). Pathways retaining fewer than 2 genes are excluded in the gene coverage filtering step. The distributions are identical across all five TCGA cohorts, as they share the same gene expression features.}
    \label{fig:gene_coverage}
\end{figure}

Each base pathway is intersected with the genes available in the cohort's expression matrix (Fig.~\ref{fig:gene_coverage}). Pathways retaining fewer than 2 genes after intersection are removed, as they provide insufficient context for bipartite message passing. Because all five TCGA cohorts share the same 4{,}999 gene features from the SurvPath repository, this step is identical across cohorts: 22 of the 684 base pathways (all Reactome) have fewer than 2 genes represented among the SurvPath features and are excluded, yielding the final vocabulary of 662 pathways (612 Reactome, 50 Hallmark) covering 4{,}574 unique genes.

\vspace{-20pt}

\subsubsection{Pathway Variance Filtering (Optional).} An optional variance-based filter can remove pathways whose constituent genes show minimal expression variation across patients, as such pathways are unlikely to contribute discriminative signal. Pathway activity is computed as the mean expression of constituent genes per patient, and the variance of this activity score across the cohort determines pathway informativeness. In the experiments reported in the main paper, this filter retains 100\% of pathways (i.e., it is effectively disabled), as the preceding curation steps already produce a sufficiently focused vocabulary.

\begin{figure}[h!]
    \centering
    \includegraphics[width=0.6\linewidth]{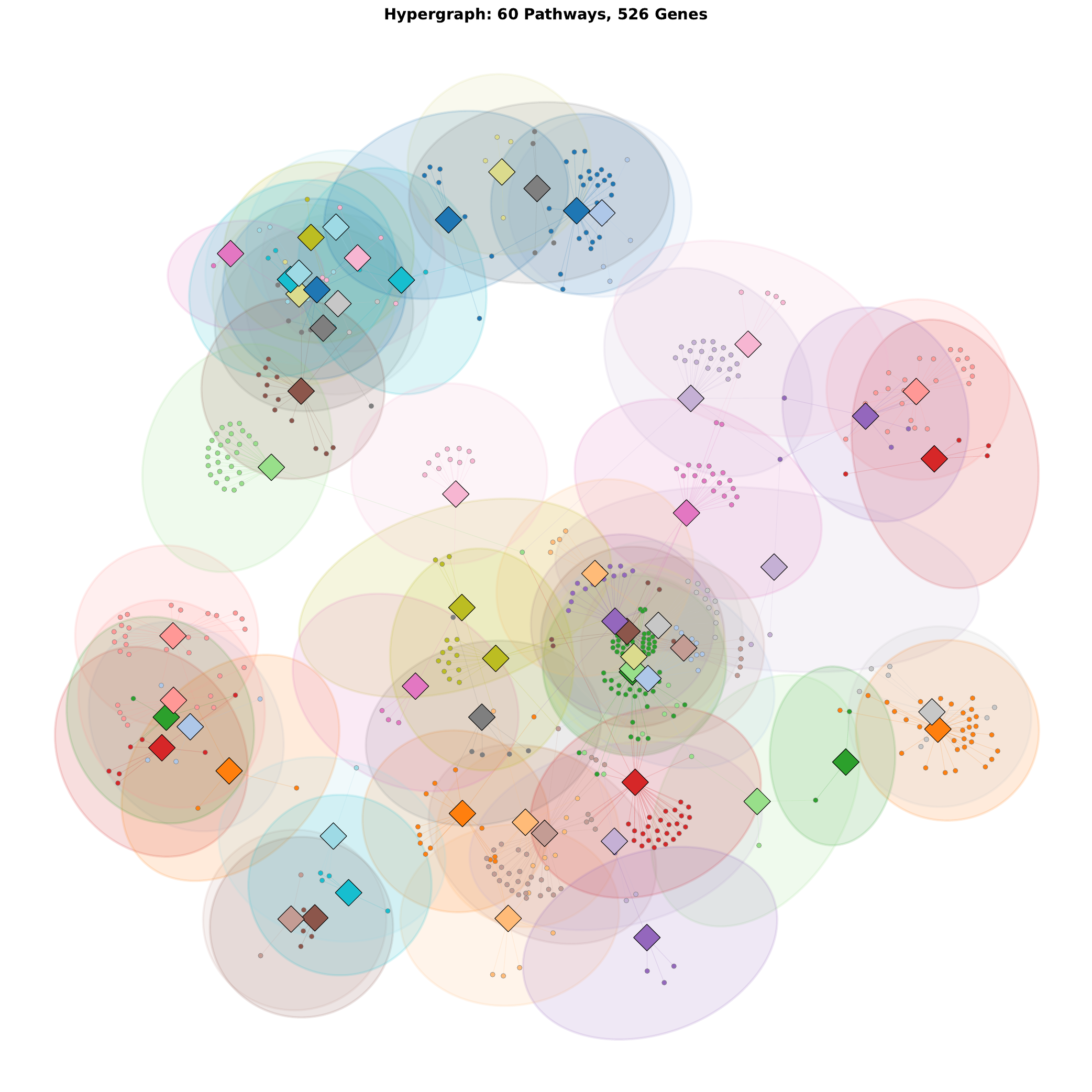}
    \caption{\textbf{Hypergraph representation of gene-pathway relationships.} A subset of 60 pathways and 526 genes displayed as a hypergraph, where each colored ellipse represents a pathway hyperedge enclosing its member genes (small dots). Central diamond nodes denote pathway identities. Overlapping ellipses indicate genes shared between pathways, reflecting the many-to-many biological relationships that the bipartite graph formulation preserves. The bipartite graph (Fig.~\ref{fig:bipartite_graph}) provides an equivalent representation with explicit pathway nodes, which is the form directly consumed by the GNN.}
    \label{fig:hypergraph}
\end{figure}

\subsubsection{Bipartite Graph Construction.} 
\begin{figure}[h!]
    \centering
    \includegraphics[width=0.6\linewidth]{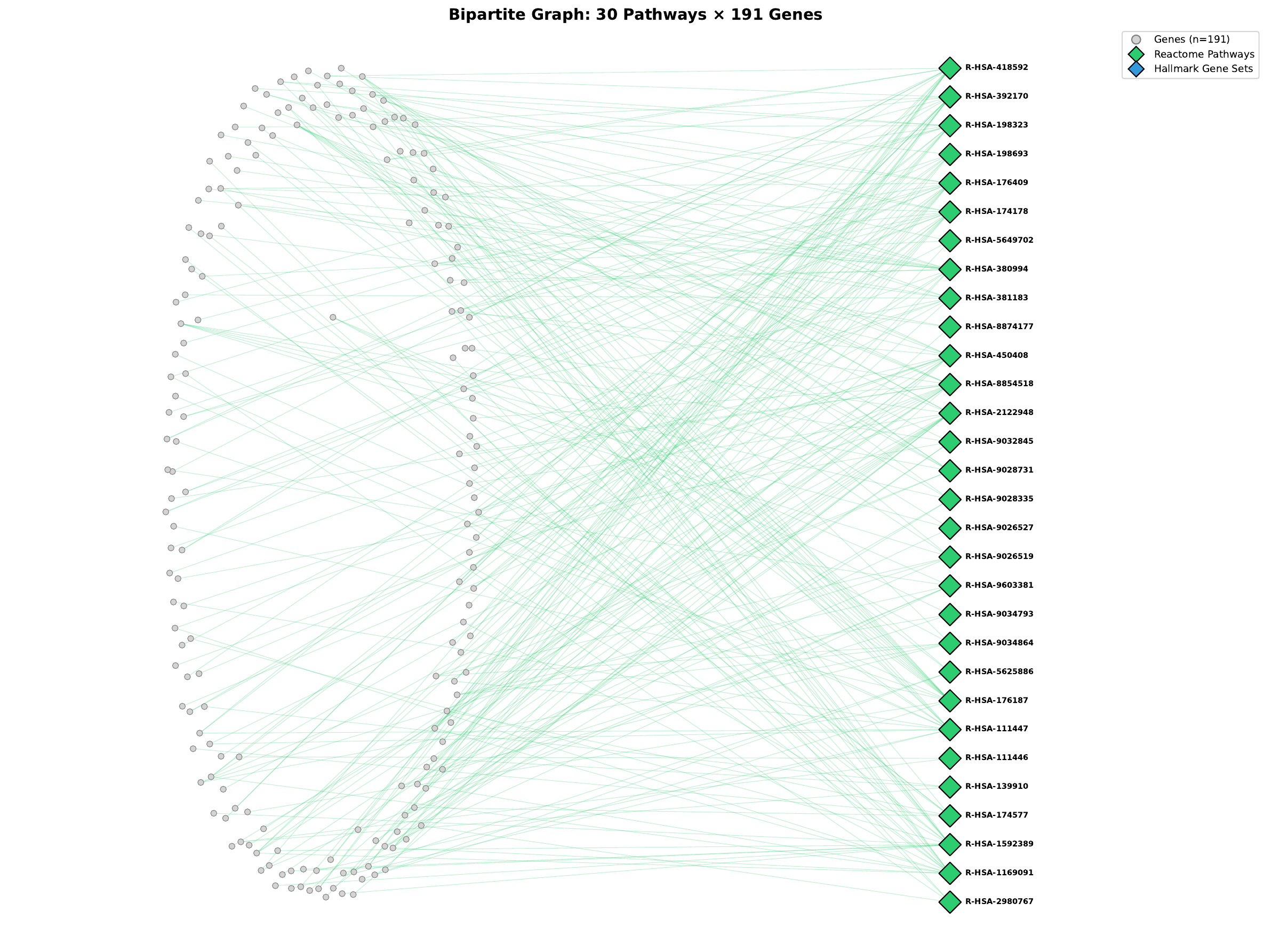}
    \caption{\textbf{Bipartite graph structure.} A subset of 30 pathways and their 191 constituent genes, visualized as a bipartite graph. Gene nodes (gray circles, left) connect via bidirectional edges to pathway nodes (green diamonds, right). Genes appearing in multiple pathways have edges to each, providing the cross-pathway communication channels exploited by message passing. The full graph comprises 662 pathways, 4{,}574 genes, and 17{,}275 bidirectional edges.}
    \label{fig:bipartite_graph}
\end{figure}

The final gene and pathway sets are assembled into the bipartite graph $\mathcal{G} = (\mathcal{V}_G \cup \mathcal{V}_P, \mathcal{E}_{GP})$ consumed by the model (Fig.~\ref{fig:bipartite_graph}). Gene nodes $\mathcal{V}_G$ correspond to genes present in both the expression data and at least one retained pathway. Pathway nodes $\mathcal{V}_P$ correspond to the retained pathways. Bidirectional edges $\mathcal{E}_{GP}$ connect each gene to every pathway it belongs to, enabling the simultaneous gene-to-pathway and pathway-to-gene message passing described in Section~2.3 of the main paper. The graph is stored as a PyTorch edge index tensor alongside mappings from node indices to gene and pathway identifiers. Figure~\ref{fig:hypergraph} provides an alternative view of the same structure as a hypergraph, where each pathway defines a hyperedge enclosing its constituent genes, illustrating the overlapping many-to-many relationships that the bipartite formulation preserves.

\subsubsection{Final Vocabulary.} Across the five TCGA cohorts used in this work, the pipeline yields 662 pathways covering 4{,}574 unique genes. The pathway size distribution is right-skewed, with most pathways containing fewer than 50 genes and a tail extending to the 200-gene upper bound, reflecting the natural hierarchy from specific molecular interactions to broader cellular programs. The vocabulary comprises predominantly Reactome pathways supplemented by the 50 MSigDB Hallmark gene sets described above. Because all cohorts share the same SurvPath gene features, the resulting bipartite graph topology is identical across datasets; only cohort size varies. Table~\ref{tab:graph_stats} summarizes the shared graph structure.

\begin{table}[h!]
    \centering
    \caption{\textbf{Bipartite graph statistics.} All five TCGA cohorts share the same 4{,}999 gene expression features, yielding an identical pathway vocabulary and graph topology.}
    \label{tab:graph_stats}
    \begin{tabular}{lr}
        \toprule
        \textbf{Statistic} & \textbf{Value} \\
        \midrule
        Input genes (SurvPath) & 4{,}999 \\
        Genes in pathways & 4{,}574 \\
        Total pathways & 662 \\
        \qquad Reactome & 612 \\
        \qquad Hallmark & 50 \\
        Bidirectional edges & 17{,}275 \\
        Mean genes per pathway & 12.8 \\
        Mean pathways per gene & 1.6 \\
        \bottomrule
    \end{tabular}
\end{table}

\subsubsection{Reproducibility.} The entire curation pipeline is deterministic and parameterized through configuration files. Stage~1 parameters (target depth, size bounds, excluded categories, Hallmark additions, Jaccard threshold) are encoded in the output filename, enabling unambiguous provenance tracking. Stage~2 records a complete preprocessing manifest including the base pathway file used, expression filtering settings, and resulting graph statistics, saved as a JSON file alongside each cohort's outputs. The pipeline can be re-executed end-to-end via two CLI commands with no manual intervention.

\section{Interpretability Signals}
\label{supp:interpretability}

ProtoPathway produces a complete set of interpretability signals during standard inference, requiring no post-hoc attribution methods. Table~\ref{tab:signals} summarizes the eight signals, here we describe their extraction in detail and explain how they compose to construct the spatial overlays shown in Fig.~\ref{fig:spatial}.

\subsection{Signal Extraction}\label{sec:sm-signal-extraction}

All signals are extracted from a single forward pass with attention logging enabled. We describe each signal below:

\subsubsection{Morphological signals (prototype encoding).} The prototype encoder (Section~\ref{sec:wsi_encoder}) produces two interpretability signals per patient, along with auxiliary quantities used in spatial overlay construction.

\medskip 

\textbf{(i) Soft assignment matrix} $\boldsymbol{\alpha} \in \mathbb{R}^{N \times K}$. For each patch $n$ and prototype $k$, the assignment probability is computed as in Eq.~\ref{eq:cosine_sim}, where the shared linear projection $f: \mathbb{R}^D \to \mathbb{R}^d$ maps both patch embeddings $\mathbf{h}_n$ and learnable prototype vectors $\mathbf{c}_k$ into a common $d$-dimensional space. Both projected vectors are $L_2$-normalized before computing cosine similarity, and the temperature parameter $\tau = 0.1$ controls assignment sharpness. The softmax is taken over the prototype dimension so that $\sum_{k=1}^K \alpha_{nk} = 1$ for every patch. From the soft assignments, two auxiliary quantities are derived: hard assignments $k^\ast_n = \operatorname{argmax}_k \alpha_{nk}$ (the discrete prototype label per patch, used for all spatial overlays), and pre-softmax cosine similarities $\mathbf{S} \in \mathbb{R}^{N \times K}$ (the raw scores before temperature scaling and softmax, used for prototype exemplar extraction in Section~\ref{sec:sm-exemplars}).

\medskip 

\textbf{(ii) Prototype gate weights} $w_k = \operatorname{softmax}_k\bigl(\psi(\mathbf{t}_k)\bigr)$, where $\psi: \mathbb{R}^d \to \mathbb{R}$ is a learned linear layer that maps each prototype token $\mathbf{t}_k$ to a scalar score, and softmax is applied across all $K$ prototypes. These weights quantify each prototype's contribution to the patient-level WSI embedding $\bar{\mathbf{z}}_\mathrm{wsi} = \sum_k w_k \mathbf{t}_k$, reflecting morphological importance before any cross-modal context is incorporated.

\subsubsection{Genomic signals (pathway encoding).} The bipartite GNN encoder (Section~\ref{sec:gene_encoder}) produces three signals:

\medskip 

\textbf{(iii) Gene-pathway attention coefficients} $\alpha_{g \to p} \in \mathbb{R}^{G \times P}$. These are the dynamic attention coefficients from the final GATv2 layer (Eq.~\ref{eq:gat}), which operates on enriched representations produced by the preceding GraphSAGE layers and LayerNorm. For multi-head attention, the per-edge coefficients are averaged across heads and reorganized from the sparse edge-indexed format into a dense $G \times P$ matrix. Because the softmax normalization is taken over each pathway's gene neighborhood independently, coefficients within a pathway sum to one: $\sum_{g \in \mathcal{N}(p)} \alpha_{g \to p} = 1$. This means they quantify relative gene importance within each pathway, not absolute importance across pathways.

\medskip 

\textbf{(iv) Pathway gate weights} $w_p = \operatorname{softmax}_p\bigl(\phi(\mathbf{z}_p)\bigr)$ (Eq.~\ref{eq:pathway_gate}), where $\phi: \mathbb{R}^d \to \mathbb{R}$ is a learned linear layer. These provide a direct measure of each pathway's prognostic contribution for the given patient.

\medskip 

\textbf{(v) Overall gene importance} $\mathbf{i}_g = \sum_{p} \alpha_{g \to p} \cdot w_p$, a derived signal that weights each gene's per-pathway attention coefficient by the corresponding pathway's importance. This provides a single scalar per gene summarizing its overall contribution to the genomic representation, accounting for both within-pathway relevance and the pathway's prognostic weight.

\subsubsection{Cross-modal signals (fusion).} The cross-attention fusion module (Section~\ref{sec:fusion}) produces three signals:

\medskip

\textbf{(vi) Cross-modal attention matrix} $\mathbf{A} \in \mathbb{R}^{K \times P}$ (Eq.~\ref{eq:attention_matrix}). Prototype tokens serve as queries and pathway embeddings serve as keys and values in multi-head scaled dot-product attention. Each entry $a_{kp}$ quantifies the association between morphological prototype $k$ and biological pathway $p$ for the given patient, with softmax normalization over pathways for each prototype: $\sum_p a_{kp} = 1$.

\medskip

\textbf{(vii) Per-prototype pathway profile} $\mathbf{A}_{k,:} \in \mathbb{R}^P$: each row of $\mathbf{A}$ provides a distribution over pathways for a single prototype, enabling pathway ranking per morphological pattern. This is a direct read-out from $\mathbf{A}$ rather than a separately computed quantity.

\medskip

\textbf{(viii) Fusion gate weights} $w_k^{\mathrm{f}} = \operatorname{softmax}_k\bigl(\psi_f(\tilde{\mathbf{t}}_k)\bigr)$ (Eq.~\ref{eq:fusion_gates}), where $\tilde{\mathbf{t}}_k$ are the cross-attended prototype representations and $\psi_f$ is a separate learned linear layer from the one used for Signal~(ii). These weights assess prototype importance after molecular context has been integrated. Comparing $w_k$ and $w_k^{\mathrm{f}}$ across the same patient reveals whether a prototype's prognostic relevance is intrinsically morphological or emerges through its association with specific molecular programs.

\subsection{Spatial Overlay Construction}
\label{sec:sm-spatial-overlays}

Since each patch has known tissue coordinates from the WSI preprocessing step, the per-patient signals described above can be projected back onto the slide to produce spatial visualizations. All overlays are rendered on a tissue canvas constructed from the original WSI file, at a configurable downsample factor. We generate four types of spatial overlay, each composing a different subset of the signals in Table~\ref{tab:signals}.

\subsubsection{Prototype Assignment Map}
\label{sec:sm-proto-overlay}

Each patch $n$ is colored by its hard-assigned prototype $k^\ast_n = \operatorname{argmax}_k \alpha_{nk}$, using a perceptually distinct $K$-color palette. This produces a spatial segmentation of the slide into morphological compartments learned entirely from the survival objective, without any tissue-type supervision. Figure~\ref{fig:spatial}A illustrates this overlay on a BLCA case.

\medskip

Construction is straightforward: for each patch, its tissue coordinates $(x_n, y_n)$ are mapped to canvas coordinates $(x_n / s,\, y_n / s)$ where $s$ is the downsample factor, and the tile at that position is filled with the color assigned to prototype $k^\ast_n$. The assignment map provides the spatial scaffold on which all subsequent overlays are built. Because every downstream signal (pathway associations, gene importance, fusion gating) is indexed by prototype, the assignment map defines which tissue regions each signal refers to.

\subsubsection{Pathway Overlay}
\label{sec:sm-pathway-overlay}

Each patch inherits a single pathway label from its assigned prototype, producing a categorical spatial map of molecular program identity over the tissue (Fig.~\ref{fig:spatial}B). To ensure the overlay highlights prognostically relevant associations, we use a risk-aware pathway selection procedure:

\medskip

For each prototype $k$, we consult the per-fold population-level rank analysis of the cross-modal attention rows $\mathbf{A}_{k,:}$ (described in SM~\ref{supp:statistics}). This analysis identifies, for each prototype, which pathways have the largest rank difference between model-predicted high-risk and low-risk groups. For a high-risk patient, prototype $k$ receives its top high-risk-associated pathway (largest positive rank difference); for a low-risk patient, it receives its top low-risk-associated pathway (largest negative rank difference). Each unique pathway in the resulting map is assigned a distinct color, and every patch is filled with the color of its prototype's selected pathway. This design means the pathway overlay adapts to the patient's risk direction: for the same slide, a high-risk patient's overlay highlights the molecular programs the model associates with poor prognosis, while a low-risk patient's overlay highlights those associated with favorable prognosis.

\subsubsection{Single-Pathway Heatmap}
\label{sec:sm-single-pathway}

For a selected pathway $p$, a continuous heatmap visualizes the spatial distribution of the model's attention to that pathway across the tissue (Fig.~\ref{fig:spatial}C). The per-patch value is:
\begin{equation}
v_n^{(p)} = a_{k^\ast_n,\, p},
\end{equation}
where $k^\ast_n$ is the patch's hard prototype assignment and $a_{k,p}$ is the $(k, p)$ entry of the cross-modal attention matrix $\mathbf{A}$. Each patch thus receives the attention its prototype allocates to the selected pathway. The values are rendered using the \texttt{inferno} colormap. Because cross-modal attention values are softmax-normalized over $P = 662$ pathways, raw magnitudes concentrate in a narrow range. To spread the colormap across the full dynamic range, we apply a rank transformation: per-patch values are replaced with their within-slide percentile ranks (scaled to $[0, 1]$). This produces overlays where color intensity reflects relative spatial variation in attention for the selected pathway, independent of its absolute magnitude. The rank transformation is applied as $\tilde{v}_n^{(p)} = (\operatorname{rank}(v_n^{(p)}) - 1) \,/\, (N - 1)$, where $\operatorname{rank}(\cdot)$ uses average-rank tie breaking.

\subsubsection{Single-Gene Heatmap}
\label{sec:sm-single-gene}

For a selected gene $g$, a continuous heatmap visualizes where the gene's signal is spatially concentrated across the tissue (Fig.~\ref{fig:spatial}D). This overlay composes the gene-pathway attention from the GATv2 layer with the cross-modal prototype-pathway attention, yielding a per-patch score:
\begin{equation}
v_n^{(g)} = \sum_{p=1}^{P} a_{k^\ast_n,\, p} \cdot \alpha_{g \to p},
\end{equation}
where $a_{k^\ast_n, p}$ is the cross-modal attention from the patch's prototype to pathway $p$, and $\alpha_{g \to p}$ is the GATv2 attention coefficient of gene $g$ within pathway $p$. The dot product measures how much of the gene's genomic signal flows through the morphological region occupied by each patch: it is high when the patch's prototype strongly attends to pathways in which the gene is important, and low otherwise. As with the single-pathway heatmap, a rank transformation maps the values to within-slide percentile ranks before colormap application. This is the default for publication figures and is what is shown in Fig.~\ref{fig:spatial}D.

\subsection{Prototype Exemplar Extraction}
\label{sec:sm-exemplars}

To aid morphological interpretation of the learned prototypes, we extract the most representative patches for each prototype. Patches are ranked by their pre-softmax cosine similarity to the prototype in the projected $d$-dimensional space (the auxiliary similarity scores derived from Signal~i), and the top-$M$ patches (default $M = 8$) are selected. We use the pre-softmax similarity rather than the post-softmax assignment probabilities because the latter are normalized across prototypes and therefore less discriminative for within-prototype ranking: a patch may have high assignment probability to a prototype simply because it has low similarity to all others, rather than because it is a strong exemplar. Exemplar patches are cropped from the original WSI, when selecting which prototypes to display, we rank prototypes by their gate weight and show exemplars for the top-$K'$ prototypes. Figure~\ref{fig:proto_exemplars} shows per-prototype exemplar grids for BLCA.

\subsection{Complete Attribution Chain}\label{sec:sm-attribution-chain}

The individual signals described above compose into a multi-level attribution chain linking molecular programs to spatial tissue locations, entirely from the architecture's inherent computation.

\begin{itemize}
    \item[\textbullet] At the lowest level, GATv2 attention $\alpha_{g \to p}$ (Signal~\textit{iii}) quantifies each gene's contribution within its pathway(s). Because the bipartite graph connects genes to all pathways they belong to, a single gene can have different attention coefficients in different pathways, reflecting context-dependent importance. One level up, pathway gate weights $w_p$ (Signal~\textit{iv}) quantify each pathway's contribution to the patient-level genomic embedding; together with Signal~\textit{iii}, this yields overall gene importance (Signal~\textit{v}).

    \item[\textbullet] On the cross-modal side, the attention matrix $\mathbf{A}$ (Signal~\textit{vi}) links each pathway to the morphological patterns it is associated with. Finally, patch-prototype assignments (Signals~\textit{i} and~\textit{ii}) map each prototype to specific tissue coordinates, completing the chain from genes through pathways and prototypes to spatial locations.

    \item[\textbullet] The spatial overlays described in Sections~\ref{sec:sm-proto-overlay} instantiate specific traversals of this chain. The prototype assignment map uses the final link alone. The pathway overlay composes the cross-modal attention (Signal~\textit{vi}) with the assignment map. The single-pathway heatmap isolates one column of $\mathbf{A}$ and projects it through the assignment map. The single-gene heatmap composes the gene-pathway attention (Signal~\textit{iii}), the cross-modal attention (Signal~\textit{vi}), and the assignment map, traversing the full chain from a single gene to every patch on the slide.
\end{itemize}

\section{Statistical Analysis}
\label{supp:statistics}

The interpretability signals described in SM~H characterize individual patients. To identify signals systematically associated with survival outcomes across the cohort, we employ population-level statistical analysis. This section describes the rank-based testing framework, the meta-analytic combination across cross-validation folds, and the specific analyses applied to each signal type.

\subsection{Motivation: Rank-Based Analysis}\label{sec:sm-rank-motivation}

Raw attention magnitudes from softmax-normalized mechanisms are difficult to compare across patients and across signals. Different patients have different numbers of patches, different expression profiles, and different overall attention scales, so a pathway gate weight of 0.003 in one patient is not directly comparable to 0.003 in another. More fundamentally, softmax normalization couples the magnitudes of all entries: an increase in one pathway's weight necessarily decreases others, making absolute values sensitive to the composition of the full set rather than reflecting independent importance.

\medskip

We therefore adopt a rank-based approach. For each patient, raw importance scores are converted to within-patient ranks, eliminating inter-patient variation in overall magnitude and focusing the analysis on relative prioritization: which pathways (or genes, or prototypes) does the model rank highest for this patient? The subsequent statistical tests compare the distributions of these ranks between risk groups. This approach is robust to monotone transformations of the importance scores and makes no distributional assumptions, which is appropriate for the small, heterogeneous cohorts typical of clinical survival studies. This population-level approach is also immune to the Jain and Wallace~\cite{jain} critique of individual-level attention faithfulness, because it treats attention as a distributional property of the model's behavior across a patient population rather than requiring that any single patient's attention weights faithfully represent the model's internal reasoning for that patient.

\subsection{Within-Patient Ranking}\label{sec:sm-within-patient-ranking}

For each patient $i$, the raw importance vector $\mathbf{w}_i \in \mathbb{R}^E$ (where $E$ is the number of entities, e.g., $P$ pathways or $K$ prototypes) is converted to a rank vector $\mathbf{r}_i \in \mathbb{R}^E$ using average-rank tie breaking. That is, entity $j$ receives rank $r_{ij} = \operatorname{rank}(w_{ij})$ among $\{w_{i1}, \ldots, w_{iE}\}$, with ties assigned the mean of the ranks they would occupy. Higher ranks correspond to higher importance. This transformation is applied independently to each patient before any between-group comparisons.

\subsection{Fold-Stratified Testing}\label{sec:sm-fold-stratified}

Different cross-validation fold models learn different attention patterns because they are trained on different data subsets with different initializations. Pooling patients across folds and comparing high-risk versus low-risk groups would conflate genuine risk-associated signal with inter-model variance, potentially drowning out true effects or producing false positives driven by a single fold's idiosyncrasies. We therefore conduct all statistical tests independently within each fold.

\medskip

Within each fold $k$, patients are split into high-risk and low-risk groups by median predicted risk score. For each entity $j$, a two-sided Mann-Whitney $U$ test compares the within-patient rank distributions $\{r_{ij} : i \in \text{low-risk}\}$ versus $\{r_{ij} : i \in \text{high-risk}\}$. This tests whether the entity tends to be ranked differently (higher or lower relative to the other entities) in high-risk versus low-risk patients. The test requires at least two patients per group within the fold; folds not meeting this criterion are excluded.

\subsubsection{Effect size.} The rank-biserial correlation provides a standardized effect size on $[-1, 1]$:
\begin{equation}
r = 1 - \frac{2U}{n_\mathrm{low} \cdot n_\mathrm{high}},
\end{equation}
where $U$ is the Mann-Whitney $U$ statistic and $n_\mathrm{low}$, $n_\mathrm{high}$ are the group sizes. Positive values indicate higher ranks in the high-risk group (the entity is more important for high-risk predictions), and negative values indicate higher ranks in the low-risk group.

\subsubsection{Rank difference.} For interpretive display (e.g., bar charts in Figs.~\ref{fig:proto_importance} and~\ref{fig:pathway_genes}), we also report the mean rank difference $\bar{r}_{\mathrm{high}} - \bar{r}_{\mathrm{low}}$ for each entity. This is on the scale of rank positions and provides an intuitive measure of how many positions higher or lower the entity is ranked in high-risk versus low-risk patients.

\subsection{Meta-Analysis Across Folds}\label{sec:sm-stouffer}

Per-fold $p$-values and effect sizes are combined across the $F$ cross-validation folds using Stouffer's weighted $Z$ method. For each entity $j$:

\begin{enumerate}
\item Each per-fold $p$-value $p_{jk}$ is converted to a directional $Z$-score: $z_{jk} = \Phi^{-1}(1 - p_{jk}/2) \cdot \operatorname{sign}(r_{jk})$, where $\Phi^{-1}$ is the standard normal quantile function and $r_{jk}$ is the rank-biserial correlation for entity $j$ in fold $k$. The sign ensures that consistent directional effects across folds reinforce each other, while inconsistent directions cancel. The $p$-value is clamped to $[10^{-15},\, 1 - 10^{-15}]$ before inversion for numerical stability.

\item Fold weights are set proportional to the square root of the fold's total patient count: $w_k = \sqrt{n_k}$, giving more influence to larger folds.

\item The combined $Z$-score is:
\begin{equation}
Z_j = \frac{\sum_{k=1}^{F} w_k \, z_{jk}}{\sqrt{\sum_{k=1}^{F} w_k^2}},
\end{equation}
and the combined $p$-value is $p_j = 2\,(1 - \Phi(|Z_j|))$.

\item The combined effect size is the weighted mean: $\bar{r}_j = \sum_k w_k \, r_{jk} \,/\, \sum_k w_k$.
\end{enumerate}

An entity requires results from at least two folds for meaningful combination. Multiple testing across all entities is corrected via the Benjamini-Hochberg procedure at $\alpha = 0.05$.

\subsection{Application to Specific Signals}\label{sec:sm-signal-applications}

The fold-stratified analysis pipeline is applied to the following interpretability signals:

\begin{itemize}

    \item[\textbullet] \textbf{Pathway gate importance (Signal \textit{iv}).} The entity vector for each patient is $\{w_p\}_{p=1}^P$, the pathway gate weights from the genomic encoder. This analysis identifies pathways systematically prioritized by the model in high-risk versus low-risk predictions. Pathway identities are consistent across folds (all folds share the same Reactome vocabulary), so Stouffer's combination applies directly. Results are shown in Fig.~\ref{eq:pathway_gate}A.

    \item[\textbullet] \textbf{Gene importance (Signal \textit{v}).} The entity vector for each patient is the overall gene importance $\{\mathbf{i}_g\}_{g=1}^G$, computed as the pathway-weighted sum of GATv2 attention coefficients. We compute two variants: the sum $\mathbf{i}_g = \sum_p \alpha_{g \to p} \cdot w_p$ and the average $\mathbf{i}_g = (\sum_p \alpha_{g \to p}) \,/\, |\{p : \alpha_{g \to p} > 0\}|$, the latter normalizing for the number of pathways a gene participates in. Both are combined across folds via Stouffer's method.

    \item[\textbullet] \textbf{Within-pathway gene attention (Signal \textit{iii}).} For selected pathways of interest (typically the top pathways from the pathway gate analysis), the entity vector is restricted to genes belonging to that pathway: $\{\alpha_{g \to p}\}_{g \in \mathcal{N}(p)}$. This identifies which genes drive a specific pathway's importance. Gene identities are consistent across folds, so Stouffer's combination applies. Results for the top high-risk and low-risk pathways are shown in Fig.~\ref{eq:pathway_gate} (A1 and A2).

    \item[\textbullet] \textbf{Prototype gate importance (Signals \textit{ii} and \textit{viii}).} The entity vectors are the WSI gate weights $\{w_k\}_{k=1}^K$ and fusion gate weights $\{w_k^{\mathrm{f}}\}_{k=1}^K$, respectively. Because prototypes are re-initialized from $k$-means centroids per fold, prototype identities are not directly comparable across folds: ``Prototype 0'' in fold 0 may correspond to a different morphological concept than ``Prototype 0'' in fold 1. The rank analysis for these signals is therefore performed at the individual fold level rather than combined via Stouffer's method, and results (Fig.~\ref{fig:proto_importance}) are reported per fold.

    \item[\textbullet] \subsubsection{Cross-modal attention per prototype (Signal \textit{vii}).} For each prototype $k$, the entity vector is the row $\mathbf{A}_{k,:} \in \mathbb{R}^P$ of the cross-modal attention matrix. This identifies which pathways each morphological pattern is most strongly associated with in each risk direction. As with prototype gate importance, this analysis is performed per fold due to prototype re-initialization across folds. The per-prototype pathway rankings are used by the pathway overlay construction (SM~\ref{supp:interpretability}) and by the cross-modal heatmap summaries.
    
\end{itemize}

\subsection{Gating Shift Analysis}
\label{sec:sm-gating-shift}

To quantify how cross-modal context reshapes morphological priorities, we compare prototype importance before fusion (WSI gate $w_k$, Signal~ii) and after fusion (fusion gate $w_k^{\mathrm{f}}$, Signal~viii) across the patient population. For each patient, both weight vectors are independently converted to within-patient ranks over the $K$ prototypes. The rank shift for prototype $k$ is defined as the change in rank position from pre-fusion to post-fusion: $\Delta_k = \operatorname{rank}(w_k^{\mathrm{f}}) - \operatorname{rank}(w_k)$. A positive shift indicates that the prototype gained importance after pathway context was incorporated; a negative shift indicates it lost importance. Because both gate vectors are observed for the same patient, this is a within-patient paired comparison that does not require cross-fold combination.

\medskip

The population-level shift is computed as the mean $\Delta_k$ across all patients within a fold. To assess whether the shifts differ by risk group, we compare the distributions of $\Delta_k$ between high-risk and low-risk patients for each prototype. Results are visualized as a three-panel figure (Fig.~\ref{fig:proto_importance}): pre-fusion importance by risk group (left), the mean rank shift between gates (center), and post-fusion importance by risk group (right).

\section{Hyperparameters}
\label{supp:hyperparams}

Table~\ref{tab:hyperparams} summarizes the hyperparameters used for ProtoPathway. The majority of architectural and optimization settings are shared across all five cohorts: both encoders project to a common dimensionality of $d{=}128$, the gene encoder uses three GNN layers, the WSI encoder uses $K{=}16$ prototypes with softmax temperature $\tau{=}0.1$, and all models are trained with AdamW at a weight decay of $10^{-5}$ for up to 100 epochs with early stopping on validation concordance index.

A small number of hyperparameters are tuned per cohort via validation performance within the five-fold cross-validation protocol. These are limited to regularization strength (dropout in the gene encoder and fusion module), the number of attention heads in the gene encoder and fusion module, and the encoder learning rate. In practice, the per-cohort settings fall into two configurations: BLCA and STAD use dropout 0.5 with four attention heads in both the gene encoder and fusion module, while BRCA, COADREAD, and HNSC use dropout 0.25 with two fusion attention heads. Encoder learning rates range from $1 \times 10^{-4}$ to $2 \times 10^{-4}$. All methods in the comparison use the same predefined folds and evaluation procedure.

\begin{table}[h]
\centering
\caption{ProtoPathway hyperparameters. Top: shared across all cohorts. Bottom: per-cohort settings. All models are trained with AdamW, batch size 1, and no learning rate scheduler.}
\label{tab:hyperparams}
\small
\resizebox{0.4\linewidth}{!}{%
\begin{tabular}{llc}
\toprule
\textbf{Component} & \textbf{Hyperparameter} & \textbf{Value} \\
\midrule
Gene encoder & Hidden dim ($d$) & 128 \\
             & Num.\ GNN layers & 3 \\
WSI encoder  & Hidden dim ($d$) & 128 \\
             & Num.\ prototypes ($K$) & 16 \\
             & Temperature ($\tau$) & 0.1 \\
Fusion       & Hidden dim & 128 \\
Training     & Weight decay (L2) & $10^{-5}$ \\
             & Max epochs & 100 \\
             & Survival bins & 4 \\
             & Seed & 42 \\
\bottomrule
\end{tabular}%
}

\vspace{0.5em}

\resizebox{0.9\linewidth}{!}{%
\begin{tabular}{llccccc}
\toprule
\textbf{Component} & \textbf{Hyperparameter} & \textbf{BRCA} & \textbf{BLCA} & \textbf{COADREAD} & \textbf{HNSC} & \textbf{STAD} \\
\midrule
Gene encoder & Dropout & 0.25 & 0.5 & 0.25 & 0.25 & 0.5 \\
             & Attention heads & 2 & 4 & 4 & 4 & 4 \\
Encoder      & Learning rate & $2 \times 10^{-4}$ & $1 \times 10^{-4}$ & $1 \times 10^{-4}$ & $1 \times 10^{-4}$ & $2 \times 10^{-4}$ \\
Fusion       & Attention heads & 2 & 4 & 2 & 2 & 4 \\
             & Dropout & 0.25 & 0.5 & 0.25 & 0.25 & 0.5 \\
\bottomrule
\end{tabular}%
}
\end{table}

\section{Datasets}
\label{supp:datasets}

We evaluate ProtoPathway on five cancer cohorts from The Cancer Genome Atlas (TCGA)~\cite{tcga}, summarized in Table~\ref{tab:dataset_stats}. Cohort selection follows SurvPath~\cite{survpath} to enable direct comparison with prior multimodal survival methods.

\paragraph{Cohort descriptions.}

\begin{itemize}

    \item[\textbullet] TCGA-BRCA (Breast Invasive Carcinoma) is the largest cohort, comprising predominantly early-stage breast cancers with a correspondingly low event rate. The high censoring fraction reflects the generally favorable prognosis of breast cancer relative to the other cohorts included here.

    \item[\textbullet] TCGA-BLCA (Bladder Urothelial Carcinoma) consists of muscle-invasive bladder cancers.

    \item[\textbullet] TCGA-COADREAD combines colon adenocarcinoma (COAD) and rectum adenocarcinoma (READ) into a single colorectal cohort, following standard practice in multimodal survival benchmarking~\cite{survpath}.

    \item[\textbullet] TCGA-HNSC (Head and Neck Squamous Cell Carcinoma) includes cancers of the oral cavity, oropharynx, larynx, and hypopharynx, representing a heterogeneous anatomical distribution.

    \item[\textbullet] TCGA-STAD (Stomach Adenocarcinoma) captures gastric cancers with diverse molecular subtypes.
    
\end{itemize}

\paragraph{Survival endpoint.}
We use overall survival (OS) as the clinical endpoint, defined as the time from diagnosis to death by any cause. Censored patients (those alive at last follow-up or lost to follow-up) are right-censored. Continuous survival times are discretized into $B = 4$ quantile-based bins computed from uncensored patients, following Zadeh and Schmid~\cite{zadeh}. The event rate (proportion of uncensored patients) varies substantially across cohorts, as shown in Table~\ref{tab:dataset_stats} from approximately 15\% in BRCA to approximately 50\% in BLCA and STAD, reflecting the differing natural histories of these cancers.

\begin{table}[t]
\centering
\caption{Dataset statistics for the five TCGA cohorts used in this study. $N$ denotes the number of patients with both WSI and gene expression data available in the predefined splits. Event rate is the proportion of uncensored (deceased) patients.}
\label{tab:dataset_stats}
\small
\resizebox{0.95\linewidth}{!}{%
\begin{tabular}{llccc}
\toprule
\textbf{Cohort} & \textbf{\quad Cancer Type} & $N$ & \textbf{Event Rate} & \textbf{Folds} \\
\midrule
TCGA-BRCA & \quad Breast Invasive Carcinoma & 714 & 12.5\% & 5 \\
TCGA-BLCA & \quad Bladder Urothelial Carcinoma & 359 & 43.5\% & 5 \\
TCGA-COADREAD & \quad Colorectal Adenocarcinoma & 227 & 21.1\% & 5 \\
TCGA-HNSC & \quad Head \& Neck Squamous Cell Carcinoma & 392 & 44.9\% & 5 \\
TCGA-STAD & \quad Stomach Adenocarcinoma & 318 & 37.7\% & 5 \\
\midrule
\multicolumn{2}{l}{\textit{Total}} & 2{,}010 & & \\
\bottomrule
\end{tabular}
}
\end{table}

\section{Ablation Studies}
\label{supp:ablation}

We ablate three architectural choices: modality branches (Section~\ref{supp:unimodal_ablation}), fusion mechanism (Section~\ref{supp:fusion_ablation}), and prototype count (Section~\ref{supp:k_ablation}). In each case, the design was motivated by interpretability: the ablations verify that these choices do not sacrifice predictive performance.

\subsection{Unimodal Ablation}
\label{supp:unimodal_ablation}

The unimodal variants reported in Table~\ref{tab:tcga_cindex} (\textsc{ProtoPath}$_\text{wsi}$ and \textsc{ProtoPath}$_\text{gene}$) are trained from scratch with one branch architecturally removed, not post-hoc evaluations of the multimodal model. When a branch is disabled via configuration (\texttt{model.branches.gene} or \texttt{model.branches.wsi}), the corresponding encoder is not instantiated, the fusion module is removed, and the classifier is replaced with a fresh linear layer mapping from $d$ directly to the $B$ survival bin logits. Training follows the same folds, survival objective, early stopping criterion, and hyperparameters as the multimodal setting. Prototype centroids follow the same $k$-means initialization described in SM~\ref{supp:kmeans}.

\subsection{Fusion Ablation}
\label{supp:fusion_ablation}

Table~\ref{tab:fusion_ablation} compares ProtoPathway's cross-attention fusion against three alternatives within the same architecture, with all other hyperparameters held fixed. ProtoPathway achieves the highest concordance index on every cohort and the highest overall C-index (0.670), outperforming concatenation (0.614), bilinear (0.630), and gated fusion (0.630). The advantage is largest on HNSC (+0.060 over gated, the next best on that cohort), where cross-modal integration appears most critical to surpassing gene-only baselines. Concatenation performs worst overall, consistent with its inability to model interactions between modality streams: it simply stacks the two pooled embeddings and relies on the downstream MLP to discover any cross-modal structure. Bilinear and gated fusion offer modest improvements by introducing multiplicative interactions, but neither matches cross-attention, which explicitly queries each pathway embedding from each prototype token.

\begin{table}[h]
\centering
\caption{Fusion ablation study. Concordance index across five TCGA cohorts for each fusion variant, with all other hyperparameters held fixed. Best performance in \textbf{bold}.}
\label{tab:fusion_ablation}
\small
\resizebox{0.95\linewidth}{!}{%
\begin{tabular}{lccccc c}
\toprule
\textbf{Fusion} & \textbf{BRCA} & \textbf{BLCA} & \textbf{COADREAD} & \textbf{HNSC} & \textbf{STAD} & \textbf{Overall} \\
\midrule
Concatenation   & 0.586$\pm$0.036 & 0.627$\pm$0.050 & 0.689$\pm$0.144 & 0.549$\pm$0.068 & 0.617$\pm$0.051 & 0.614 \\
Bilinear        & 0.584$\pm$0.047 & 0.635$\pm$0.055 & 0.719$\pm$0.131 & 0.547$\pm$0.053 &  0.664$\pm$0.055 & 0.630 \\
Gated           & 0.589$\pm$0.047 & 0.636$\pm$0.048 & 0.701$\pm$0.132 & 0.582$\pm$0.078 &  0.627$\pm$0.066 & 0.630 \\
\midrule
ProtoPathway & \textbf{0.649$\pm$0.050} & \textbf{0.646$\pm$0.075} & \textbf{0.740$\pm$0.132} & \textbf{0.642$\pm$0.047} & \textbf{0.674$\pm$0.069} & \textbf{0.670} \\
\bottomrule
\end{tabular}%
}
\end{table}

Beyond predictive performance, the cross-attention formulation is the only variant that produces interpretable cross-modal associations: the $K \times P$ attention matrix $\mathbf{A}$ provides a per-patient mapping from morphological prototypes to biological pathways, enabling the spatial overlays and population-level analyses described in SM~\ref{supp:interpretability} and SM~\ref{supp:statistics}. The alternative fusion mechanisms aggregate the two modalities into a single vector without exposing which pathways each prototype attends to, precluding the complete attribution chain that is central to ProtoPathway's design. The ablation therefore confirms that the interpretability-motivated choice of cross-attention does not come at the cost of predictive performance, it is also the strongest fusion strategy on this benchmark.

\subsection{Prototype Count ($K$)}
\label{supp:k_ablation}

The number of prototypes $K$ controls the granularity of the morphological decomposition: too few prototypes may conflate distinct tissue compartments, while too many may fragment coherent patterns into redundant or noisy clusters. We evaluate $K \in \{4, 8, 16, 32, 64\}$ on BLCA, with all other hyperparameters are held fixed at the values in Table~\ref{tab:hyperparams}, and prototypes are re-initialized via $k$-means for each $K$ following the procedure in SM~\ref{supp:kmeans}.

\begin{figure}[h]
    \centering
    \includegraphics[width=0.65\linewidth]{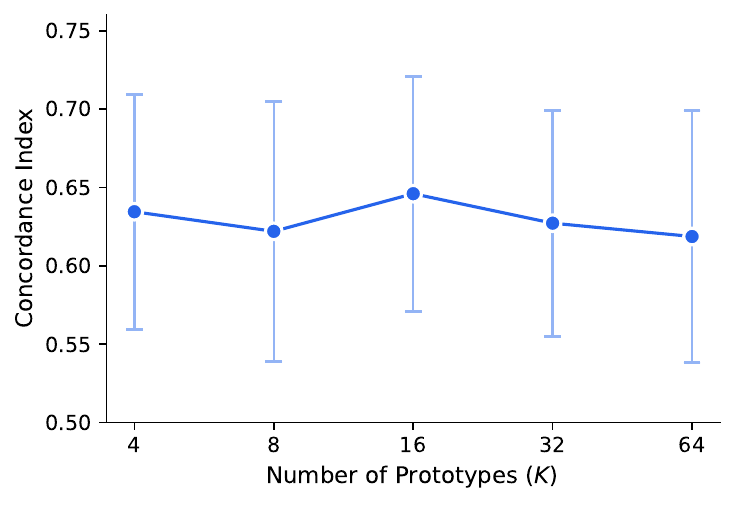}
    \caption{Sensitivity of ProtoPathway to the number of WSI prototypes $K$ on TCGA-BLCA. Error bars denote standard error across five cross-validation folds. Performance is stable across the tested range, with $K{=}16$ achieving the highest concordance index.}
    \label{fig:k_sensitivity}
\end{figure}

Figure~\ref{fig:k_sensitivity} reports the results. Performance is broadly stable across the tested range, with all configurations falling within overlapping standard error intervals, indicating that the model is not highly sensitive to this hyperparameter. $K{=}16$ achieves the highest C-index (0.646), while both smaller and larger values show modest decreases. At $K{=}4$, the model remains competitive (0.635) but compresses BLCA's diverse tissue landscape into too few compartments to support fine-grained spatial interpretation. At $K{=}64$, performance drops to 0.619, suggesting that an excess of prototypes fragments coherent tissue compartments into redundant clusters, diluting the gating signal without adding discriminative capacity.
 
The choice of $K{=}16$ balances predictive performance with interpretive resolution. At this granularity, the learned prototypes recover clinically meaningful tissue compartments (SM~\ref{supp:supp_proto}) without excessive fragmentation: each prototype aggregates a sufficient number of patches to produce a stable token embedding, and the $16 \times 662$ cross-modal attention matrix remains compact enough for direct inspection. Smaller values risk merging tissue types that carry distinct prognostic signals (e.g., collapsing muscle and connective tissue into a single prototype), while larger values split coherent compartments and produce prototypes that are difficult to interpret morphologically.

\section{Model Efficiency}
\label{supp:efficiency}

Table~\ref{tab:efficiency} compares the computational cost of ProtoPathway against all baselines, profiled on 30 randomly selected patients from TCGA-STAD. Among multimodal methods, ProtoPathway achieves the fastest training time at 13.6\,ms per patient, a 28--50$\times$ speedup over the cross-attention and masked self-attention baselines (MCAT at 380\,ms, MOTCAT at 534\,ms, SurvPath at 476\,ms, PIBD at 685\,ms, MMP at 432\,ms). The only multimodal method with comparable speed is PORPOISE (11.7\,ms), which employs a simpler Kronecker product fusion without cross-modal attention. Inference times follow the same pattern, with ProtoPathway at 4.7\,ms compared to 189--377\,ms for the attention-based multimodal baselines.

Critically, this efficiency does not come at the cost of predictive performance. As shown in Table~\ref{tab:tcga_cindex}, ProtoPathway achieves the highest overall concordance index (0.670) across the five TCGA cohorts, ranking first on four of five datasets and second on the remaining one. The methods closest in overall performance, MCAT (0.662) and SurvPath (0.660), require 28$\times$ and 35$\times$ longer to train per patient respectively, and consume comparable or greater VRAM (289\,MB and 818\,MB vs.\ 325\,MB). PORPOISE, the only multimodal method faster than ProtoPathway, achieves a substantially lower overall concordance index (0.650) with a 3$\times$ larger parameter count (1.51M vs.\ 480K).

The efficiency advantage stems from the prototype bottleneck in the WSI encoder. Standard cross-attention between gene tokens and the full patch set scales quadratically with the number of patches, which for typical whole slide images ranges into the thousands. ProtoPathway instead compresses the variable-length patch set into $K{=}16$ fixed prototype tokens via soft assignment before any cross-modal interaction occurs. The cross-attention in the fusion module therefore operates over 16 tokens rather than the full patch sequence, reducing both FLOPs and memory. This is reflected in the VRAM footprint: ProtoPathway requires 325\,MB compared to 818\,MB for SurvPath, 948\,MB for PIBD, and 1817\,MB for MMP.

\begin{table}[h!]
\centering
\caption{Computational efficiency comparison across models, run on 30 randomly selected patients from TCGA-STAD (validation fold 0). Params $=$ trainable parameters. FLOPs estimated per patient (single forward pass). VRAM $=$ peak GPU memory during training. Train/Infer $=$ time per patient, excluding data loading.}
\label{tab:efficiency}
\resizebox{0.9\linewidth}{!}{%
\begin{threeparttable}
\begin{tabular}{l c c c c c c}
\toprule
Model & Modality & Params & FLOPs & VRAM (MB) & Train (ms) & Infer (ms) \\
\midrule
  MLP & Gene & 783.4K & 1.6M & 68 & 3.0 & 0.5 \\
  SNN & Gene & 782.3K & 85.3M & 68 & 2.7 & 0.5 \\
\midrule
  ABMIL & WSI & 399.7K & 4.82G & 70 & 2.7 & 0.7 \\
  DSMIL & WSI & 2.37M & 28.95G & 170 & 7.5 & 2.7 \\
  TransMIL & WSI & 2.94M & 77.27G & 865 & 43.5 & 13.0 \\
\midrule
  PORPOISE & MM & 1.51M & 11.21G & 164 & 11.7 & 1.6 \\
  MOTCAT & MM & 611.0K & 4.18G & 287 & 533.6 & 306.5 \\
  MCAT & MM & 677.0K & 5.59G & 289 & 380.4 & 188.5 \\
  SurvPath & MM & 473.6K & 29.19G & 818 & 475.7 & 262.3 \\
  PIBD & MM & 1.25M & $-\tnote{1}$ & 948 & 684.7 & 377.4 \\
  MMP & MM & 427.0K & 779.7M & 1817 & 431.5 & 208.6 \\
\textbf{ProtoPathway} & MM & 479.7K & 3.86G & 325 & 13.6 & 4.7 \\
\bottomrule
\end{tabular}
\begin{tablenotes}
\item [1] \footnotesize{we were unable to compute FLOPs for PIBD due to auxiliary loss computation involving stochastic sampling and information bottleneck operations that break PyTorch's tracing}
\end{tablenotes}
\end{threeparttable}}
\end{table}

The parameter count of ProtoPathway (480K) is comparable to the most compact multimodal baselines (MMP at 427K, SurvPath at 474K) and substantially smaller than PORPOISE (1.51M) or PIBD (1.25M). The FLOPs (3.86G) are moderate, falling below PORPOISE (11.21G) and SurvPath (29.19G), though above MMP (780M). MMP achieves lower FLOPs through GMM-based prototype assignment but at considerably higher VRAM (1817\,MB) and wall-clock time (432\,ms), suggesting memory-intensive operations not captured by forward-pass FLOPs alone. This dissociation underscores that FLOPs are an incomplete proxy for practical efficiency in multimodal survival models, and that ProtoPathway's prototype bottleneck provides a more favorable trade-off across all four efficiency dimensions: parameters, FLOPs, VRAM, and wall-clock time.


\section{Prototype Morphology}
\label{supp:supp_proto}

\begin{figure}
    \centering
    \includegraphics[width=0.7\linewidth]{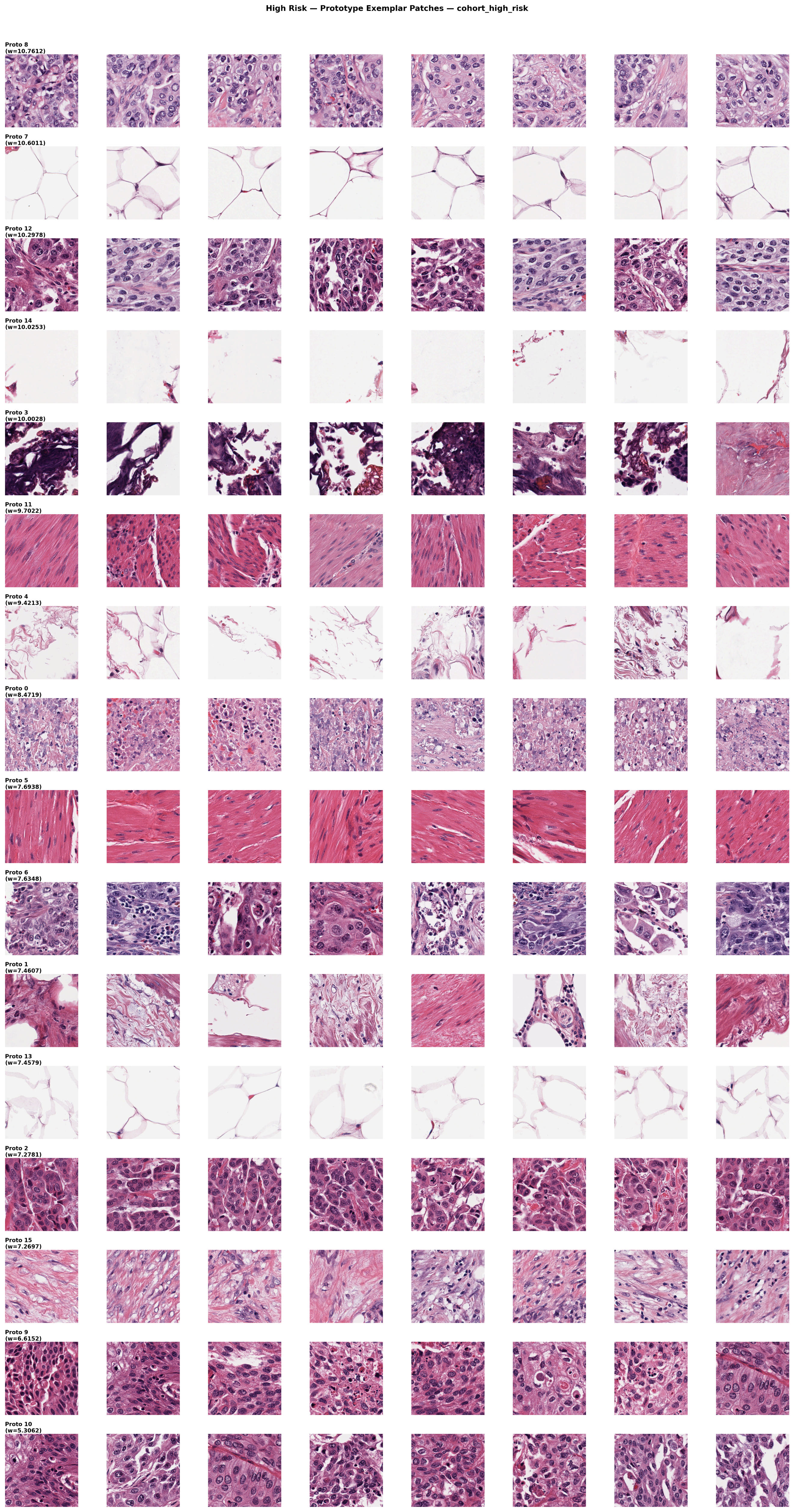}
    \caption{\textbf{Prototype exemplar patches for BLCA (Fold~1).} For each of the $K{=}16$ learned prototypes, the eight patches with highest cosine similarity to the prototype centroid in the projected space are shown. Prototypes are ordered by descending gate weight $w_k$ (shown in parentheses). No tissue-type labels are used during training; the morphological groupings emerge entirely from the survival objective.}
    \label{fig:proto_exemplars}
\end{figure}

Figure~\ref{fig:proto_exemplars} presents the eight nearest-neighbor exemplar patches for each of the 16 learned prototypes in the BLCA cohort, ordered by descending gate importance. Without explicit tissue-type supervision, the learned prototypes recover a morphological vocabulary that partitions the tissue landscape into coherent compartments: multiple tumor phenotypes (solid, nested, glandular, and immune-infiltrated); two muscle states (intact bundles versus loosely arranged fibers); two adipose variants (densely and loosely packed perivesical fat); graded connective tissue responses from loose, vascularized stroma to dense fibrosis; and necrotic debris. This decomposition aligns with clinically meaningful staging boundaries in bladder cancer: connective tissue invasion defines pT1 disease, muscle involvement defines $\geq$pT2, and fat invasion defines $\geq$pT3a. The model also separates tumor prototypes by immune infiltration density, a feature increasingly relevant to treatment selection~\cite{babjuk2022european}. Below, we describe each prototype in detail:

\begin{itemize}
    \item[\textbullet] \textbf{Proto~8 - Dense tumor cells.} Tightly packed, darkly stained cancer cells filling the field of view with minimal surrounding tissue. Cells are large and irregular, arranged in confluent sheets without clear structure.
    
    \item[\textbullet] \textbf{Proto~7 - Fat tissue (perivesical).} Large, round, empty-appearing cells characteristic of adipose (fat) tissue. Consistently pure across all exemplars. Cancer presence in this layer indicates advanced ($\geq$pT3a) disease.
    
    \item[\textbullet] \textbf{Proto~12 - Tumor nests in surrounding stroma.} Discrete clusters of cancer cells embedded in reactive connective tissue. Unlike Proto~8, cells form distinct islands rather than confluent sheets, indicating a more structured growth pattern.

    \item[\textbullet] \textbf{Proto~14 - Background / catch-all prototype.} Predominantly empty patches with minimal tissue content, including slide background and tissue preparation artefacts.

    \item[\textbullet] \textbf{Proto~3 - Dead tissue and inflammatory debris.} Dark, fragmented tissue with cellular debris and dense immune cell infiltration. Several exemplars show regions of cell death with faint outlines of former cells, a hallmark of aggressive tumor behavior.

    \item[\textbullet] \textbf{Proto~11 - Intact muscle bundles.} Thick, elongated pink fibers arranged in tight parallel bundles. The well-preserved, orderly architecture and bundle thickness indicate muscularis propria (the bladder's main muscle wall) rather than thinner superficial muscle layers.

    \item[\textbullet] \textbf{Proto~4 - Sparse, fibrosed connective tissue.} Pale, sparsely cellular tissue with thin strands of collagen and scattered blood vessels. Consistent with the lamina propria layer in a fibrosed or swollen state---the tissue layer relevant to early-stage (pT1) invasion.

    \item[\textbullet] \textbf{Proto~0 - Tumor invasion front.} Irregular cancer cell clusters infiltrating into dense, reactive connective tissue. The interface between tumor and surrounding stroma is clearly visible, with some exemplars showing isolated cells or small groups at the leading edge of invasion.

    \item[\textbullet] \textbf{Proto~5 -  Loosely arranged smooth muscle.} Elongated spindle-shaped nuclei within thick pink fibers, less tightly bundled than Proto~11, consistent with muscularis propria sectioned obliquely or with reactive stromal changes.

    \item[\textbullet] \textbf{Proto~6 - Tumor with gland-like structures.} Irregular cancer cell clusters with attempted tube or cavity formation, mixed with immune and connective tissue cells. This architectural variation from the solid pattern of Proto~8 is clinically significant, as non-standard growth patterns carry independent prognostic value in bladder cancer.

    \item[\textbullet] \textbf{Proto~1 - Loose, vessel-rich connective tissue.} Pale, loosely organized connective tissue with visible thin-walled blood vessels and scattered immune cells. Less fibrosed than Proto~4, representing a more viable variant of the submucosal connective tissue layer.

    \item[\textbullet] \textbf{Proto~13 - Loosely packed adipose tissue.} Large, well-separated adipocytes with thin intervening septa. Distinguished from Proto~7 by larger cell size and sparser packing, consistent with the outermost perivesical fat layer where adipocytes are less compressed by surrounding structures.

    \item[\textbullet] \textbf{Proto~2 - Tumor with lymphocytic infiltration.} Dense clusters of large, irregular tumor cells intermixed with prominent populations of small, dark lymphocytes. Distinguished from the solid tumor of Proto~8 by the conspicuous immune cell component, and from Proto~15 by a higher tumor-to-lymphocyte ratio. The degree of immune infiltration within the tumor compartment is increasingly recognized as prognostically and therapeutically relevant in bladder cancer.

    \item[\textbullet] \textbf{Proto~15 - Immune-cell-rich tumor.} Cancer cells intermixed with dense populations of small, dark immune cells (tumor-infiltrating lymphocytes). The prominent immune presence distinguishes this from the solid (Proto~8) and nested (Proto~12) tumor prototypes, and is relevant to predicting response to immunotherapy.

    \item[\textbullet] \textbf{Proto~9 - High-grade, disorganized tumor.} Dense sheets of highly irregular cells with chaotic architecture. More disorganized than Proto~8, with less cell-to-cell cohesion and features suggestive of non-standard differentiation patterns.

    \item[\textbullet] \textbf{Proto~10 - Tumor nests in reactive stroma.} Irregular clusters of tumor cells embedded in cellular, reactive connective tissue. Similar to Proto~12 but with smaller, less well-defined nests and a more disorganized tumor--stroma interface, suggesting a less structured invasive growth pattern.

\end{itemize}

\end{document}